%% file: 00-main.tex
\documentclass[preprint, review, 3p, authoryear, times]{elsarticle}

\makeatletter
\def\ps@pprintTitle{%
 \let\@oddhead\@empty
 \let\@evenhead\@empty
 \let\@oddfoot\@empty
 \let\@evenfoot\@empty
}
\newcommand{\rev}[1]{\textcolor{black}{#1}}
\makeatother
\input{Content/01-packages}

\begin{document}

\input{Content/02-title-authors_es}
\input{Content/03-abstract}
\input{Content/10-intro}
\input{Content/20-related-work}

\input{Content/30-applications}

\input{Content/40-future-discussion}

\input{Content/50-conclusion}

\input{Content/60-acknowledgements}

\small
\bibliographystyle{elsarticle-harv}
\bibliography{99-references}

\end{document}

%% file: Content/01-packages.tex
\usepackage{graphicx}
\usepackage{url}
\usepackage{color, soul}
\usepackage{bm}
\usepackage{amsmath} 
\usepackage{lineno}

\usepackage{geometry}
\usepackage{array}

\usepackage{longtable}
\usepackage{booktabs}
\usepackage{url,lineno,microtype,subcaption, graphicx,float}
\usepackage[table]{xcolor}

%% file: Content/02-title-authors_es.tex
\title{\rev{Leveraging Generative AI for Urban Digital Twins: A Scoping Review on the Autonomous Generation of Urban Data, Scenarios, Designs, and 3D City Models for Smart City Advancement}}


\author{Haowen Xu} 
\ead{xuh4@ornl.gov}
\author{Femi Omitaomu\corref{cor1}}\ead{omitaomuoa@ornl.gov}
\author{Soheil Sabri} \ead{soheil.sabri@ucf.edu}
\author{Sisi Zlatanova} \ead{s.zlatanova@unsw.edu.au}
\author{Xiao Li} \ead{xiao.li@ouce.ox.ac.uk}
\author{Yongze Song} \ead{Yongze.Song@curtin.edu.au}

\cortext[cor1]{Corresponding author.}

\address[CSED]{Computational Urban Sciences Group, Oak Ridge National Laboratory, Oak Ridge, TN 37830, USA}
\address[UCF]{Urban Digital Twin Lab, School of Modeling, Simulation, and Training, University of Central Florida , 3100 Technology Parkway,  Orlando, 32826 FL, USA}
\address[UNSE]{GRID, School of Built Environment, UNSW Sydney, NSW 2052 Australia}
\address[Oxford]{Transport Studies Unit, University of Oxford, South Parks Road, Oxford, OX1 3QY, UK}
\address[Curtin]{School of Design and the Built Environment, Curtin University, Kent St, Bentley WA, AU}


\makeatletter
\newcommand{\printfnsymbol}[1]{%
  \textsuperscript{\@fnsymbol{#1}}%
}

\newcommand*{\MyIndent}{\hspace*{0.5cm}}%


%% file: Content/03-abstract.tex
\begin{abstract} \label{sec:abstract}
The digital transformation of modern cities by integrating advanced information, communication, and computing technologies has marked the epoch of data-driven smart city applications for efficient and sustainable urban management. Despite their effectiveness, these applications often rely on massive amounts of high-dimensional and multi-domain data for monitoring and characterizing different urban sub-systems, presenting challenges in application areas that are limited by data quality and availability, as well as costly efforts for generating urban scenarios and design alternatives. As an emerging research area in deep learning, \rev{Generative Artificial Intelligence (GenAI)} models have demonstrated their unique values in content generation. This paper aims to explore the innovative integration of \rev{GenAI} techniques and urban digital twins to address challenges in the \rev{planning and management of built environments with focuses on various urban sub-systems, such as transportation, energy, water, and building and infrastructure}. The survey starts with the introduction of \rev{cutting-edge} generative AI models, \rev{such as the Generative Adversarial Networks (GAN), Variational Autoencoders (VAEs), Generative Pre-trained Transformer (GPT)}, followed by a \rev{scoping} review of the existing urban science applications that leverage the intelligent and autonomous capability of these techniques to facilitate  \rev{ the research, operations, and management of critical urban subsystems, as well as the holistic planning and design of the built environment.} Based on the review, \rev{we} discuss potential opportunities and technical strategies that integrate \rev{GenAI} models into the next-generation urban digital twins for more intelligent, scalable, and automated \rev{smart city development and management}.  

\end{abstract}

\maketitle

%% file: Content/10-intro.tex
\section{Introduction} \label{sec:Introduction}
\rev{Planning and managing urban environments often require complex, interdisciplinary efforts that encompass various urban subsystems, including transportation and mobility, urban energy, water supply and distribution, as well as buildings and infrastructure \citep{fedorowicz2020leveraging}. With the recent advent of the Internet of Things (IoT), Information and Communication Technologies (ICT), and 5G cellular networks, cities worldwide have been digitally transformed to generate substantial amounts of urban data, opportunities for more intelligent and efficient smart city applications to enhance urban sustainability, productivity, and livability \citep{hossain2018edge}. This transformation also has reshaped the management of the various urban subsystems through the integration of Artificial Intelligence (AI) and emerging digital twin technologies, which, in combination, created a new paradigm for developing urban digital twins, which is also known as digital twin cities, city digital twins, and smart city digital twins, to promote smart city services \citep{shahat2021city, dembski2020urban}. }

\rev{The original digital twin concept focuses on the development of virtual models of real-world objects and processes in a computing environment to support advanced manufacture and analytics \citep{ivanov2020digital}, and was initially proposed to support advanced manufacturing. Similarly, the urban digital twin paradigm replicates different urban subsystems in the digital environment to enable a variety of advanced capabilities, including real-time situational awareness and predictive analytics of complex urban dynamics, as well as the informed decision support, optimization, and automation of sophisticated urban management tasks through the integration of advanced visualization and visual analytics \citep{beckett2022smart, xu2022towards}, and cyber-physical integration \citep{allam2019big,luckey2021artificial}, and City Information Models (CIMs) \citep{shi2023ontology}}.

\subsection{\rev{Digital Twins for Smart Cities}  \label{sec:udt-overview}}
\rev{Following this paradigm, numerous operational urban digital twins have been developed and implemented to support the management of various real-world urban subsystems, including urban mobility \citep{xu2023smart}, building operations \citep{bortolini2022digital}, urban energy and power grids \citep{cioara2021overview,jafari2023review}, and infrastructure \citep{callcut2021digital,sabri2023designing}. In the broader context of urban planning and design, digital twins offer a holistic view of the urban ecosystem, enabling planners to create more livable, resilient, and adaptive urban spaces \citep{dembski2020urban}. Examples include the use of urban digital twins for water-sensitive urban design \citep{langenheim2022adapting} and generative development envelope control analysis \citep{sabri2022innovative}. The incorporation of urban digital twins in these sectors not only streamlines planning processes but also facilitates the examination of scenarios and the development of innovative solutions to contemporary urban challenges, paving the way for smarter, more sustainable, human-centered, and participatory cities. }

\rev{By examining contemporary digital twin studies, We have created Table \ref{tab:udt} to provide an overview of previous successful urban digital twin applications developed in various urban sectors. The table utilizes major application areas defined by recent review articles on digital twins for various urban subsystems \citep{ivanov2020digital}, including transportation \citep{bao2021review,wang2022mobility}, urban energy \citep{jafari2023review,pan2020digital}, urban water \citep{ramos2023digital}, and building and infrastructure \citep{al2021digital}. We also explored the utilization of digital twin on the planning, design, and management of the urban environment as the system of multiple subsystems from a holistic aspect \citep{al2021digital}.
}

\input{Content/Table/table-0}

\subsection{Challenges in Urban Digital Twins}  \label{sec:udt-challenges}
Although there have been significant advancements in AI, communication, and computing technologies, urban digital twins, \rev{driven by the big data,} still encounter various methodological and technical challenges in many aspects. \rev{We have outlined 6 major challenges based on the urban digital twin articles reviewed in the previous subsection. These challenges are labeled from C1 to C5 in the following list. }
\begin{description}
   \item[C1. Data Availability and Quality:] Among the massive smart city data, only a small fraction is utilized by smart services to promote urban management \citep{mohammadi2018enabling}, as much raw data collected from IoT and smart devices, and crowed-sourcing platforms are considered as digital wastes. This is because most of the raw urban data are either unlabeled or have limited data quality, rendering these data unusable for many urban science applications involving data and physical modeling \citep{park2021auto, alwan2022data}. Furthermore, a variety of urban data, such as building occupancy, energy consumption, human mobility, and vehicle trajectories, frequently faces limited availability issues owing to data privacy and residence concerns, heterogeneous sources, and nonuniform distribution of sensors \citep{figueira2022survey, zheng2014urban}. Subsequently, these datasets exhibit varying coverage, accuracy, and resolution levels across different spatial regions, rendering them less applicable to many analytical applications. 
   
   \item[C2. Hypothetical Urban Scenarios:] Many smart city applications involve strategic approaches that envision and analyze various potential future scenarios to enhance urban planning and management \citep{perveen2017evaluating, sokolov2019scenario} in different sectors, such as traffic \citep{feng2023trafficgen, xinxin2020csg}, building energy \citep{ruan2023optimization,pippia2021scenario}, and water management \citep{herman2016synthetic, mikovits2018importance}. These approaches often require costly, time-consuming, and labor-intensive efforts for collecting scenario-specific data, as well as for generating the corresponding scenarios using processed data and simulations to support predictive analytics and decision-making \citep{dong2021smart, argota2022getting}. In addition, it is also extremely difficult to collect real-world data to describe a future or hypothetical scenario, which are often considered as corner cases. 
   
   \item[C3. 3D City Modeling:] There is an increasing trend for smart city initiatives to develop 3D city models 
   as critical components of urban digital twins to support data integration and visualization, interactive planning and simulation, real-time monitoring and response, and public engagement \rev{\citep{deng2021systematic,biljecki2015applications}}. However, developing a comprehensive 3D city model for large urban areas is a costly, repetitive, and labor-intensive endeavor \citep{botin2022digital,singh2021digital}, which involves clarification of needed urban entities (e.g., buildings, transportation infrastructure, green areas, water bodies) the levels of details, the geometric accuracy, the type representations (vector or voxel), the validity specifications (e.g.  volmetric objects have to be water-tight), the involvement of photo-realistic texture (as in reality models), the data structure for management of the 3D models as well as the allowed data types (e.g. surfaces, volumes, B-splines, parametric shapes). Additionally, constructing valid 3D model, preventing intersections and discrepancies between terrain, underground and above ground urban objects, can also be a time-consuming effort \citep{xu2023toward}.  The integration of data for 3D city models relies on clear specifications of vocabularies, semantic definitions, explicit relationships and constrains. Despite the international standardization efforts and developed transformation and integration tools, many aspects of creating 3D City models  require manual or semi-automatic cleaning, adjustment and conversion. 
   
   \item[\rev{C4. Urban Design Scalability}:] Built upon the previous data analytics, simulations, and digital twin applications, the planning and management of smart cities often involve top-down and multi-scale urban design processes, which are responsible for the configuration of land-use and land-cover, transportation network, buildings, and infrastructure \citep{van2021introduction,perez2023methodology}. These processes are often repetitive, expensive, and time-consuming, involving multiple design alternatives and complex decision-making for design optimization. Therefore, it is challenging to scale up the design processes to large urban areas, which often have distinct design requirements (e.g., green space and accessibility). \rev{Subsequently, there is a need to developed automated generative co-design tools that can automated repeated and laborious processes. }

   \item[\rev{C5. Data Sovereignty Requirements}:] \rev{Data ownership, data colonialism, and data residency are key concerns related to data sovereignty. These issues can significantly constrain and restrict the sharing and usage of urban data in smart city applications \citep{rijshouwer2022public, rogan2019universal}. Data ownership complexities arise when multiple stakeholders, such as governments, private companies, and individuals, claim rights over the data, leading to disputes and hindering collaborative efforts \citep{lofgren2020value}. Data colonialism exacerbates these challenges by creating power imbalances where external entities exploit local data resources without equitable benefits to the local communities, fostering mistrust and resistance to data sharing \citep{leese2022data}. Data residency requirements further complicate matters, as strict regulations mandate that data must be stored and processed within specific geographic boundaries, limiting cross-border data flows and integration \citep{navarrete2009information}. These issues collectively impede the seamless exchange of urban data, essential for developing and deploying innovative smart city solutions, ultimately restricting the potential for enhanced urban management and improved quality of life for residents.}

    \item[\rev{C6. Public Participation and Community Engagement}]: \rev{Engaging citizens and residents without an urban science background in participatory and human-centered urban planning and smart city management presents a significant challenge \citep{senior2023evaluating}. The complexity of urban planning concepts and the technical details of smart city initiatives can be daunting for individuals lacking specialized knowledge and access to specialized software tools, creating a technical barrier to effective engagement \citep{chourabi2012understanding, tan2020smart}. Communication barriers further complicate this issue, as translating technical jargon into accessible and meaningful information is challenging. In this regard, it is essential to build a human-AI partnership through AI agents or chatbots that possess advanced human language understanding, domain knowledge, and the ability to autonomously perform specialized tasks such as knowledge base querying, data analysis, and simulation. These capabilities can be enabled through APIs integrated with urban digital twins. The AI agent should be able to directly communicate with various stakeholders, understand their individual needs, and formulate citizen-centric planning and management strategies. This approach ensures that the AI agents can facilitate meaningful engagement, making urban planning more inclusive and responsive to the community's diverse needs.} 
\end{description}

Given the complexities, crafting wide-ranging and insightful solutions to urban issues from an all-encompassing viewpoint is proving to be progressively more expensive and demanding in terms of effort. These solutions require the analysis of complex urban systems — including transportation, energy, building, infrastructure, and water — as interconnected subsystems at the ground, above-ground, and underground levels \citep{gallotti2021complex}. \rev{The selection of urban subsystems discussed in this paper is guided by previous literature and urban digital twin applications \citep{ferre2022adoption,hamalainen2021urban}, with the topic further refined by examining successful urban digital twin implementations detained in Section \ref{sec:udt-overview}}. Consequently, there is a growing need for developing scalable, autonomous, \rev{and intelligent} solutions. Such solutions would employ GenAI techniques to enhance the availability and quality of urban data, automate the generation of various scenarios, support the development of 3D city models, and automate complex urban design and optimization processes \citep{wang2023towards,kirwan2020smart}. \rev{Some GenAI models excel in understanding human languages and handling text, providing opportunities to create responsive AI agents that facilitate communication and co-design in planning and management. These AI agents enhance community participation and engagement by interpreting and generating natural language, making it easier for stakeholders to contribute their ideas and feedback in the planning process \citep{fares2023role}. This leads to more inclusive, transparent, and collaborative decision-making, ultimately resulting in urban designs that better reflect the needs and preferences of the community. }

\subsection{Motivation} \label{sec:motivation}
In the realm of deep learning, \rev{Generative Artificial Intelligence (GenAI)} has emerged as a groundbreaking field, showcasing its unique ability to create novel content from realistic data. This capability is increasingly relevant for enhancing smart city applications. Despite the rapid proliferation of \rev{GenAI} in various urban science and management sectors, there remains a notable absence of a thorough review encompassing mainstreaming \rev{GenAI} techniques and their contributions to urban and environmental management within the context of developing smart cities through digital twin models. 

This survey paper aims to fill this gap by providing a structured overview of current \rev{GenAI} models and their applications in diverse urban management sectors, particularly those not yet explored from the \rev{urban} digital twin perspective. Drawing upon recent literature and proposed agendas, the focus of this survey is primarily on key smart city application areas \rev{that include} transportation, energy system, urban water, and building and infrastructure \citep{shahat2021city, dembski2020urban}. 

\subsection{Review Method}  
\rev{Given the interdisciplinary nature of this paper's objectives, which encompass diverse topics from AI, urban informatics, and the built environment, we employed the scoping review methodology \citep{mak2022steps} to systematically map emerging concepts and identify research gaps and applications related to the synergy between GenAI models and urban digital twin applications in the smart city sector. This methodology provides a knowledge synthesis that is particularly useful for several specific scenarios involving emerging research areas, interdisciplinary topics, broad research questions, and complex or heterogeneous literature \citep{levac2010scoping}. To formulate this survey and identify relevant literature, we follow the scoping review methodological framework outlined by \cite{arksey2005scoping} through a series of procedures described in the following subsections. }

\subsubsection{\rev{Preliminary Search} \label{sec:preliminary}}
\rev{It is important to define clear and broad research questions that will guide the selection of the domain, scope, and focus areas of the literature review. Before formulating these questions, \cite{mak2022steps} suggests conducting a preliminary search of the literature to assess the feasibility of the review with regard to the following aspects:}
\rev{
\begin{enumerate}
    \item The breadth of the research question.
    \item Whether a scoping review on the topic has already been conducted.
    \item If there is sufficient literature to warrant a scoping review.
\end{enumerate}
}
\rev{
Based on these aspects, we conducted a preliminary search. Based on our topic and research motivation described in Section \ref{sec:motivation}, the breadth of our research scope encompasses a variety of domains and disciplines. The concept and applications of urban digital twins, also known as city digital twins, pertain to the fields of urban informatics, urban computing, city information modeling (CIM), and computational urban science \citep{shi2022prospective, deng2021systematic}. Meanwhile, GenAI algorithms and applications represent an emerging area within the field of deep learning.}

\rev{To determining whether a scoping review on this topic has already been conducted, we created a list of predefined syntax terms and conducted a literature search across major academic databases, including Google Scholar, Scopus, and IEEE Xplore. Our predefined search terms included ‘Generative Artificial Intelligence’, ‘Generative AI’, ‘GAI’, ‘GenAI’, ‘Generative’, ‘Generative Deep Learning’, ‘Digital Twin Cities’, and ‘City Digital Twins’. The search was performed on paper topics, including article titles, abstracts, and keywords. This search did not identify any previous review or survey articles on the subject.
}

\rev{To verify the quantity and availability of relevant literature, we developed a set of predefined search syntaxes tailored to the parent disciplines of 'Digital Twin Cities' and their synonyms. These parent disciplines include 'Smart Cities,' 'City Information Modeling,' 'Urban Computing,' and 'Urban Informatics,' as identified by previous studies \citep{zheng2014urban, xia2022study}. We combined these predefined syntaxes with GenAI search terms to conduct preliminary searches on the Scopus and IEEE Xplore academic databases. The Scopus search returned over 150 articles, while IEEE Xplore yielded 400 articles. Our preliminary searches confirmed that there is sufficient literature to justify proceeding with the proposed scoping review. For the subsequent in-depth literature review on individual GenAI applications, we conducted multiple searches across the two databases using refined search syntax that specifies not only the generic GenAI and smart city concepts but also detailed types of GenAI models and specialized urban research and management topics.
}
\rev{\subsubsection{Identifying the Research Questions}\label{sec:research-questions} }
\rev{Based on the results of the preliminary search, we have formulated 4 research questions and conducted the necessary preliminary searches to seek their answers. The answers to these fundamental research questions will guide the direction and structure of this scoping paper in a step-by-step manner.}
\rev{
\begin{description}
\item[Q1: How urban digital twins are applied to facilitate smart city advancement?] Based on previous reviews in urban science and urban planning, urban digital twins are developed to enhance the operation, management, and research of various urban subsystems, including transportation, energy, water, buildings, and infrastructure \citep{lu2021applications, pesantez2022using, broo2022design}. They also play a crucial role in supporting critical urban planning tasks \citep{schrotter2020digital, ivanov2020digital}.
\item[Q2: What GenAI technologies should we focus on in this review?] 
According to contemporary data science and AI research, we primarily focus on the applications of four types of generative models: Generative Adversarial Networks (GANs), Variational Autoencoders (VAEs), Generative Pre-trained Transformers (GPT), and Generative Diffusion Models \citep{bandi2023power, eigenschink2021deep}. Section \ref{sec:Generative-AI} offers detailed reviews of these models, including their history and rationale, from a data science perspective. 
We do not focus on the application of deep learning models such as Convolutional Neural Network (CNN), Neural Autoregressive Models, and LSTMs, which have content generation capabilities but are not considered genAI models by nature.
\item[Q3: How GenAI technologies are applied in the smart city sector?] 
Through our preliminary search conducted in Section \ref{sec:udt-challenges}, we did not identify any contemporary research explicitly proposing the integration of GenAI models into urban digital twins to facilitate smart city management, operation, and research. In this regard, this paper aims to outline the scope and strategies for potential GenAI-digital twin integration by investigating the data-driven nature, functionality, and challenges faced by modern urban digital twins. Centered around the challenges of urban digital twins identified in Section \ref{sec:udt-challenges}, we will review existing GenAI applications in the broader smart city sector that (1) target key urban subsystems identified by previous studies, and (2) are driven by big urban data. Our literature review in Section \ref{sec:Dataset} will provides answers to this question. 
\item[Q4: How GenAI technologies can benefit urban digital twins?] 
To address this question, a comprehensive review of GenAI applications in various urban sectors is required and is systematically presented in Section \ref{sec:Dataset} of this paper. Urban digital twins, inherently data-driven applications, have emerged alongside advancements in urban computing and informatics. The common ground between GenAI models and urban digital twins lies in their shared capability to generate and handle large volumes and varieties of urban data. Consequently, our literature review in Section \ref{sec:Dataset} focuses on aspects related to data enhancement and generation, simulation, and design. The findings and proposed vision for future opportunities arising from the integration of GenAI and digital twins are discussed in detail in Section \ref{sec:Future Discussion}.
\end{description}
}
\rev{The preliminary answers to our questions are used to formulate the research criteria for identifying relevant studies and to shape the direction and focus of this paper. Given the extensive and interdisciplinary nature of our topic, which involves both research and technical implementation across multiple domains, we justify our approach of using the scoping review methodology. This approach aims to define the fundamental scope of the emerging application areas that bridge GenAI models with urban digital twin technologies.  
}

\begin{figure*}[htb]
 \centering
\includegraphics[width=\textwidth]{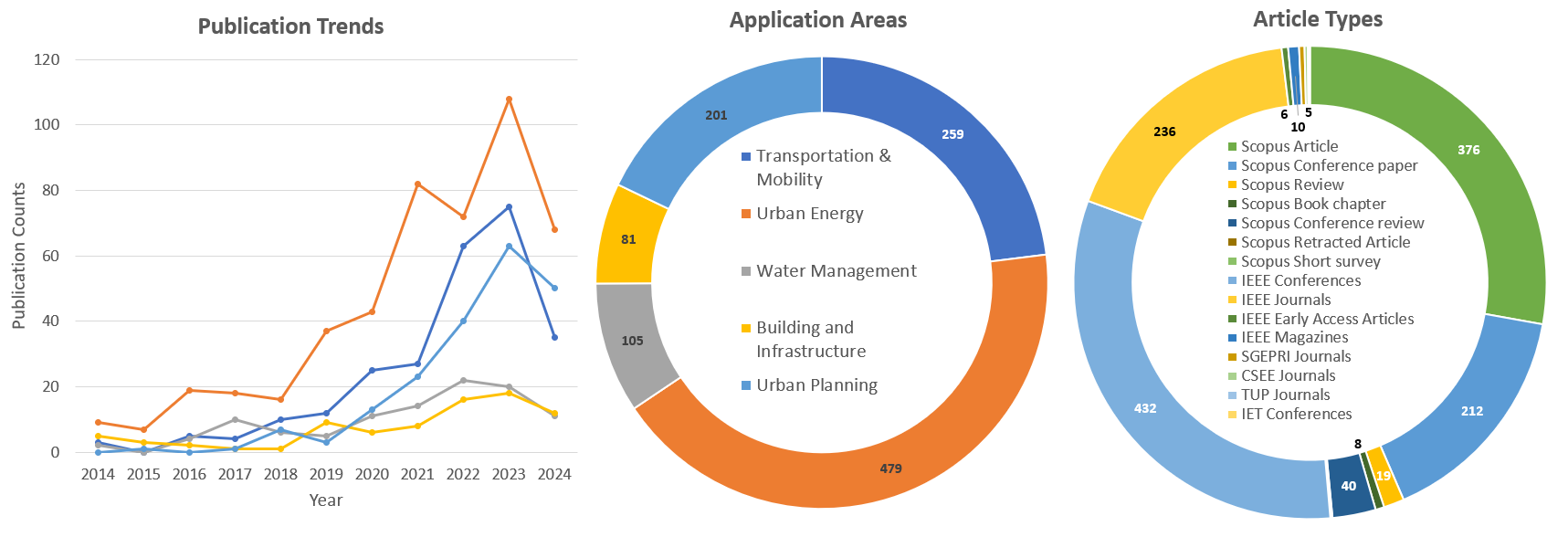}
 \caption{A summary of GenAI-related publications in major urban sectors using Scopus and IEEE Xplore database (Search conducted on July 20th 2024) }
 \label{fig:scopus-statistics}
\end{figure*}

\subsubsection{\rev{Identifying Relevant Studies} }
\rev{Through a series of standard procedures, in combined with the preliminary answers to our research questions in Section \ref{sec:research-questions}, we devise our strategy for identifying relevant studies as the following: }
\rev{
\begin{enumerate}
\item \textbf{Identify relevant databases}: Based on the subject matter, we use Scopus and IEEE Xplore as the primary academic database for conducting our search. The search is conducted based upon keywords and search syntax that matches the paper topics (including article titles, abstracts, and keywords).
\item \textbf{Define keywords and search syntax for formulating queries}: We create two categories of search keywords based on our scopes. These categories describe key concepts in both the GenAI sector and the urban science and smart city sector. For the GenAI category, keywords and search syntax are defined using vocabularies and taxonomies that describe generic GenAI concepts (e.g., ‘Generative AI’, ‘GAI’, ‘GenAI’, ‘Generative Deep Learning’) and specific GenAI models (‘GAN’, ‘VAE’, ‘GPT’, ‘LLM’, ‘Diffusion Model’). We also include various variants, abbreviated forms, and synonyms of these keywords in our search syntax. The urban science category includes keywords that define detailed application areas within individual urban subsystems, as well as generic urban planning tasks. Examples of these keywords include ‘Transportation’, ‘Mobility’, ‘Smart Grid’, ‘Building’, ‘Urban Infrastructure’, and ‘Urban Planning’. The definitions of these keywords are inspired by the urban digital twin application areas identified in Table \ref{tab:udt}.
\item \textbf{Define Inclusion and Exclusion Criteria}: The inclusion criteria prioritize the review of both peer-reviewed journal articles, conference proceedings, and government technical reports from the last 10 years. For scoping purposes, we will briefly discuss ongoing research published as preprints on SSRN and arXiv to outline potential future research directions, but we will not provide in-depth reviews or discussions of these articles in our review section. We exclude non-peer-reviewed articles, online web pages, studies outside the scope of urban research, and publications older than 10 years. Additionally, we exclude studies that focus on the cybersecurity, building construction, structure design, and agriculture aspects of urban digital twins and IoT applications, which, although relevant to smart city advancements, are outside the scope of urban research.
\end{enumerate}
}
\rev{We perform comprehensive searches based on these criteria in the selected databases. The search results are exported into a CSV file produced by reference management software (Mendeley) for charting purposes. An additional search filter includes limiting the publication language to English.
}
\begin{figure*}[htbp]
 \centering
\includegraphics[width=\textwidth]{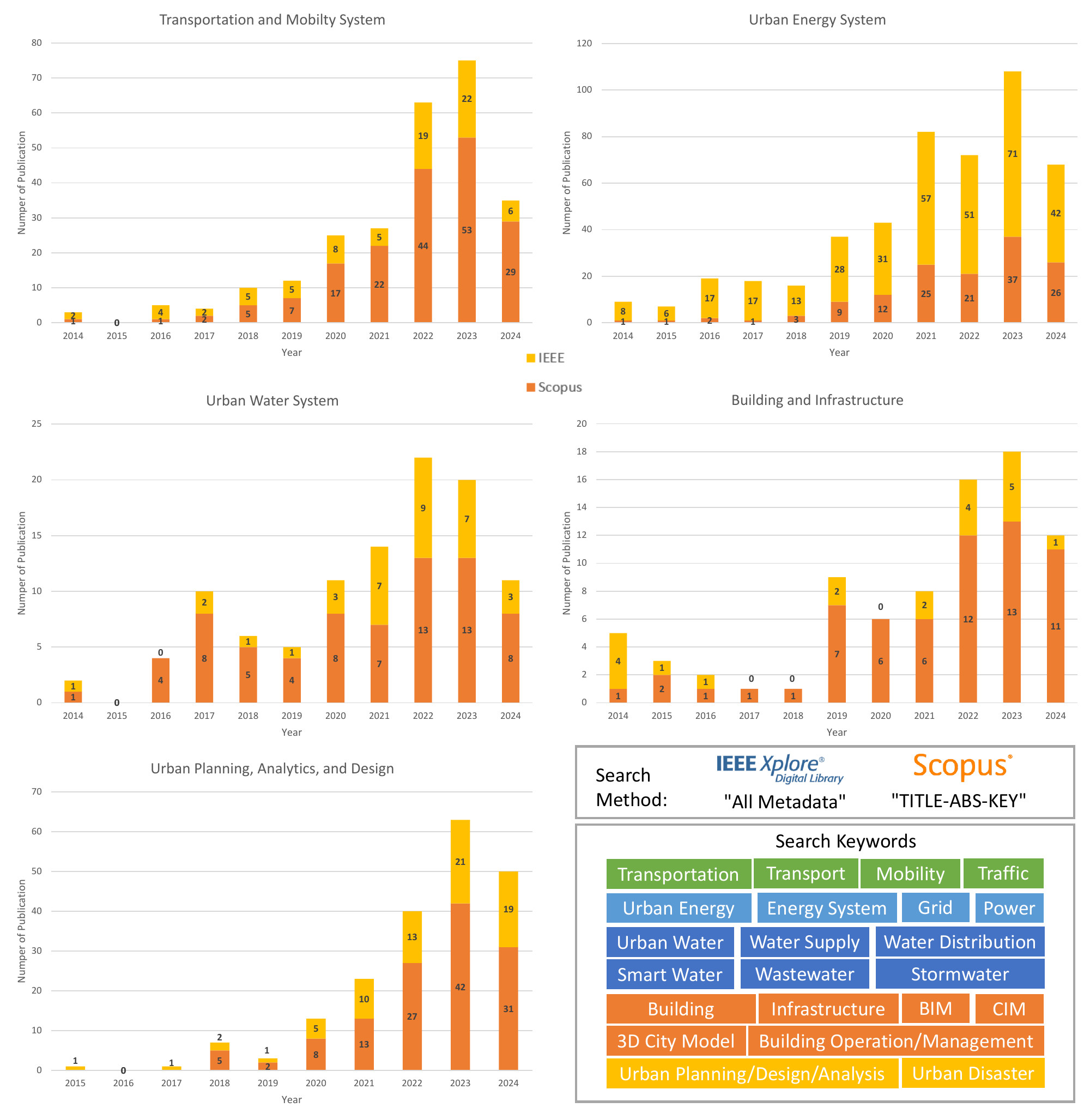}
 \caption{A summary of GenAI-related publications in major urban sectors using \rev{Scopus and IEEE xplore database (Search conducted on July 20th 2024)} }
 \label{fig:scopus-statistics2}
\end{figure*}

\subsubsection{\rev{Selecting Studies to be Included in the Review}}
\rev{We devised our search keywords based on popular types of GenAI models defined in Section \ref{sec:Generative-AI}, as well as urban application areas summarized in Table \ref{tab:udt}. These keywords were combined with logical operators and the search templates of Scopus and IEEE Xplore to formulate queries. These keywords are combined with logical operators and the search templates of Scopus and IEEE Xplore to formulate quires. With the specified quires, the initial search results yielded a total of 1,391 papers from two databases, covering 5 different urban application areas. The distribution was as follows: 298 papers in transportation and mobility, 512 in urban energy, 125 in urban water, 184 in building and infrastructure, and 272 in urban planning. After removing duplicates and excluding papers unrelated to urban science (e.g., those focused on cybersecurity and rural environmental issues) by scanning journal types, article types, abstracts, and keywords, the total number of selected papers was reduced to 1,125. The revised distribution was 259 papers in transportation and mobility, 479 in urban energy, 105 in urban water, 81 in building and infrastructure, and 201 in urban planning. These remaining papers will undergo further filtering based on human expert review, assisted by Natural Language Processing (NLP) powered Python scripts and pre-trained Large Language Models (LLMs). The NLP and LLM-based tools used to assist in literature screening extend the methodology developed by \citet{tupayachi2024towards} through the LLM-powered generation of ontology and knowledge graphs using scientific literature.
The final selected literature is summarized in Figure \ref{fig:scopus-statistics}. The article summary for individual urban application areas is provided through Figure \ref{fig:scopus-statistics2}.}

\subsubsection{\rev{Literature Review Structure}}
\rev{
The scoping review approach was chosen for its ability to provide a comprehensive overview of a broad and complex research area, identify key concepts, and map the available evidence \citep{munn2018systematic}. This method allowed us to include a wide range of study designs and sources, ensuring a thorough exploration of the topic. Our search strategy involved a systematic search of multiple databases, grey literature, and reference lists, adhering to predefined inclusion and exclusion criteria. By using the scoping review method, we aimed to identify research gaps, highlight the diversity of GenAI applications in the urban planning and smart city management sectors, and provide valuable insights for future research directions regarding their integration into urban digital twins.}

\rev{
We will begin by reviewing the five major types of GenAI models used to create our search keywords. This review will cover their rationale, capabilities, and popular applications in computer science, and is presented in Section \ref{sec:Generative-AI}. Following this, we will engage in a comprehensive discussion on GenAI applications in various urban areas, utilizing the screened literature we have collected from two academic databases. Given our focus on urban digital twins, we will further select research articles within individual urban application areas that are centered on the autonomous generation of data, scenarios, 3D city models, and designs for smart city advancement to conduct more detailed review and discussion. The in-depth scenario-based review is presented in Section \ref{sec:Dataset}.}




%% file: Content/Table/table-0.tex
\newcolumntype{M}[1]{>{\centering\arraybackslash}m{#1}}
\begin{scriptsize}
\begin{table*}[hbt]
\footnotesize
\caption{\rev{An overview of urban digital twin examples in various urban application areas.}} \label{tab:udt} 
\begin{tabular*}{\textwidth}{|p{3.4cm}|p{12.15cm}| }
 \hline
 \multicolumn{1}{c}
 {\cellcolor{black!10}\textbf{Application Area}}  & 
 \multicolumn{1}{c}{\cellcolor{black!10}\textbf{Examples}} \\
\hline         
     \begin{description}
        \item[] 
        \item[] 
        \item[] 
        \item[Transportation and Mobility] 
     \end{description}
     & 
     \begin{description}
         \item[Traffic Simulation]             \citep{kuvsic2023digital,saroj2021development, xiong2022design}  
         \item[Traffic Control Optimization] 
            \citep{xu2023smart, kamal2024digital, kumarasamy2024integration}
         \item[Traffic Data Prediction]            \citep{hu2021digital, nie2023digital,kumar2018novel}
         \item[Connected Automated Vehicles] \citep{fan2022ubiquitous,du2021digital,ali2023review, wagner2023spat}
         \item[Traffic Safety]  
            \citep{irfan2024towards,fu2024digital, lv2022digital, wang2022digital,lv2022traffic}     
         \item[Transportation Asset Management]
            \citep{gao2021digital,wu2022digital,adibfar2022creation}
        \item[Advanced Driver-Assistance Systems]
            \citep{wang2022mobility,wang2020digital, schwarz2022role}
        \item[Parking Management] \citep{chomiak2023use, zou2023digital, liu2023cognitive, shao2022computer}
     \end{description}          
      \\ 
 \hline 
     \begin{description}
        \item[]         
        \item[Urban Energy ] 
     \end{description} 
     & 
     \begin{description}
         \item[Grid Operation] \citep{jafari2023review, eGridGPT2024,sifat2023towards,nordzi2022real}
         \item[Energy Management] \citep{srinivasan2020urban, xu2024novel,francisco2020smart}
         \item[Anomaly and Fault Detection]  \citep{danilczyk2021smart, jain2019digital, joseph2018prediction}
         \item[Renewable Energy] \citep{xu2023novel, belik2023implementation, agostinelli2021cyber} 
     \end{description}
            \\     
 \hline  
    \begin{description}
        \item[]         
        \item[Urban Water] 
     \end{description} 
     & 
     \begin{description}
        \item[Smart Water Grid] \citep{wu2023high, ramos2023smart, giudicianni2020overview}        
        \item[Water Supply and Distribution]        
        \citep{conejos2020building, karmous2019foundations,martinez2021digital}
        \item[Drainage and Stormwater System] \citep{bartos2021pipedream, sharifi2024application}
        \item[Wastewater Management] \citep{hallaji2021digital, wang2024digital,jiang2022intelligent}       
     \end{description}
     \\    
 \hline
    \begin{description}
        \item[]         
        \item[Building and Infrastructure] 
     \end{description}  & 
    \begin{description}
        \item[Building Operation] \citep{zhao2022developing, hodavand2023digital, el2022development}
        \item[Building Information Modeling] \citep{qiuchen2019developing, sepasgozar2020lean, sacks2020construction}    
        \item[City Information Modeling] \citep{shi2023ontology, shariatpour2024urban, petrova2021digital}
        \item[Infrastructure Management] \citep{al2021digital, broo2022design, shirowzhan2020digital, abdeen2023citizen, yu2022digital}
     \end{description}    
    \\    
 \hline
    \begin{description}
        \item[]         
        \item[Urban Planning and Design] 
     \end{description}     
    & 
    \begin{description}
        \item[]         
        \item[] \citep{yang2021urban, schrotter2020digital, white2021digital, kumalasari2023planning, beckett2022smart, xu2023toward}
     \end{description} 
    \\
 \hline
\end{tabular*}
\end{table*}
\end{scriptsize}

%% file: Content/20-related-work.tex
\section{Generative AI Models} \label{sec:Generative-AI}
Generative AI models are a class of artificial intelligence techniques designed to generate new data samples similar to a given set of input data \citep{bandi2023power}. These models can produce a wide range of outputs, including images, text, sound, and video, and are particularly notable for their ability to create realistic, novel, and often indistinguishable data from actual human-generated content \citep{eigenschink2021deep}, allowing them to support data-driven research application in various urban science and smart city sectors. In addition, Generative AI techniques and services enabled through Large Language Models (LLMs), such as ChatGPT, have demonstrated their unique capabilities in interpreting human natural language and automating code generation and information, therefore, have been applied to develop autonomous systems in various research fields, such as geoinformatics \citep{li2023autonomous}, chemical research \citep{boiko2023autonomous}, and transportation \citep{cui2024survey}. 

\rev{Based on our search, combined with summaries proposed by other review articles \citep{lin2023generative, bandi2023power}, and the scope of GenAI models defined by \citet{bengesi2024advancements}, this paper primarily targets four GenAI models: Generative Adversarial Networks, Variational Autoencoders, Generative Pre-trained Transformers, and Generative Diffusion Models. These models and their urban applications are briefly and individually discussed in the following subsections.}

\subsection{Generative Adversarial Networks} 
\label{subsec:gan}

Generative Adversarial Networks (GANs) represent a pivotal innovation in the field of artificial intelligence, particularly within the domain of deep learning \citep{pan2019recent, goodfellow2014generative}. These models introduce a novel framework for generative modeling, leveraging the power of two neural networks—a generator and a discriminator—in a competitive setting. The generator aims to produce data indistinguishable from real data, while the discriminator endeavors to distinguish between real and generated data \citep{goodfellow2020generative, creswell2018generative}. This adversarial process results in the generator creating increasingly realistic outputs. Over time, GANs have evolved into various forms, such as Conditional GANs, which generate data based on specific conditions \citep{mirza2014conditional}, and CycleGANs, used for style transfer tasks \citep{zhu2017unpaired}. Their vast and transformative applications range from image and video generation to drug discovery and advanced image editing. GANs have revolutionized how machines understand and recreate complex data patterns, making them a cornerstone technology in modern AI research and development. Their continued evolution promises further breakthroughs in AI's capability to generate realistic content \rev{based on specific contexts}. 

Major application areas of GANs in various urban science and smart city sectors include the following: 
\begin{itemize}
  \item Image Generation: GANs can generate highly realistic images, often indistinguishable from actual photographs. This includes creating artwork, enhancing image resolution, or generating images from descriptions. In smart city applications, GANs' image generation capability is often leveraged to create synthetic urban and environmental datasets in the geospatial raster format. Examples of these synthetic data include river \citep{gautam2022realistic}, land use \citep{ansith2021modified}, and urban morphology patterns \citep{zhang2022metrogan}. 

  \item Data Augmentation: the process is used to increase the size and diversity of an existing dataset. GANs can create additional data with improved spatiotemporal resolution for training predictive models, enhancing the robustness of these models without the need to collect more real-world data. In recent years, GANs have been applied to augment urban data to improve the prediction accuracy of many urban variables in areas that include transportation management \citep{dabboussi2023gan,lin2023generative}, building energy optimization \citep{fan2022novel}, and infrastructure management \citep{gwon2023image}.
  
  \item Synthetic Data for Scenario Generation: the primary goal of synthetic data generation is to create entirely new data points that are not present in the original dataset. This is especially useful when the amount of real data available is limited (e.g., hypothetical scenarios or corner cases) or when using real data is restricted due to privacy concerns (e.g., personal or sensitive information). GANs are employed to generate synthetic data to define many hypothetical scenarios \citep{sankarscenario} with a wide range of urban management interests, such as promoting transportation planning \citep{wu2020spatiotemporal,demetriou2020generation,rempe2022generating}, urban mobility management \citep{wang2020seqst}, power system operations \citep{yuan2021multi, huang2023scenario, rizzato2022stress}, building energy modeling \citep{wang2020generating,zhang2019scenario}, and water management \citep{koochaligenerative}.
    
  \item 3D Object Generation: GANs can also generate 3D models that are useful for multiple-purpose visualization functions (e.g., virtual reality, gaming, and filmmaking) and urban science applications. These applications include creating 3D city models \citep{vesely2022building}, 3D buildings \citep{du20203d,kim2020citycraft}, and architectural design \citep{chang2021building}. The GANs' capability of generation of 3D urban objects has been employed to automate the development of 3D representation of cities in large-scale urban digital twins, which can enable virtual and augmented reality applications for scientific gamification and immersive 3D scientific visualization of urban management processes, shedding light on public engagement and participatory planning in the smart city \citep{xu2023toward}.  
\end{itemize}

\subsubsection{Conditional GAN} 
Conditional Generative Adversarial Networks (cGANs), an influential extension of the original Generative Adversarial Network (GAN) framework, were introduced to add a directed feature to the generative process \citep{mirza2014conditional}. cGANs were conceptualized to address the limitation of traditional GANs in controlling the modes of the data generation process. Mirza and Osindero (2014), in their seminal work, proposed this adaptation where both the generator and the discriminator are conditioned on some extra information, such as class labels or data from other modalities, allowing the cGANs to generate more specific and targeted outputs with improved relevance and applicability of the generated data \citep{mirza2014conditional}. The practical use cases of cGANs are vast and diverse, particularly shining in the smart city fields where tailored data generation is crucial. Examples of these fields include urban traffic estimation and predictions \citep{zhang2020curb, huang2020using}, air quality management \citep{toutouh2021conditional}, and building operations \citep{baasch2021conditional, nuastuasescu2022conditional}. The ability of cGANs to generate contextually relevant and specific data opens new possibilities in various fields, making them a focal point in ongoing AI research.


\subsubsection{CycleGANs}
CycleGAN, short for Cycle-Consistent Generative Adversarial Networks, represents a significant advancement in the field of machine learning and computer vision. This innovative framework facilitates image-to-image translation tasks without requiring paired training data \citep{zhu2017unpaired}. This is a notable departure from traditional methods that rely heavily on paired datasets, which are often difficult and expensive to obtain. CycleGAN leverages a novel approach that uses two sets of Generative Adversarial Networks (GANs) and a cycle-consistency loss function CycleGAN. This ensures that the inputs closely resemble the original inputs when translated to another domain and then back. This breakthrough has wide-ranging applications, from photo-realistic image translation to style transfer and even in tasks such as object transfiguration and season transfer in images. Its ability to learn mappings between different domains without paired examples makes CycleGAN a powerful tool in unsupervised learning, offering new possibilities in various smart city fields such as autonomous driving \citep{rajagopal2023hybrid}, augmented reality and virtual reality \citep{kim2021generating}, and context-specific data augmentation \citep{zhou2023automatic,li2021domain}.

\subsubsection{\rev{Comparison between GANs}}
\rev{
Building upon previous literature that has provided comprehensive reviews of various GAN variants and their applications in different urban science sectors, including transportation \citep{lin2023generative} and the built environment \citep{wu2022generative}, this subsection compares different GANs through Table \ref{tab:gans}. The table summarizes the differences between various types of GANs, highlighting their strengths and weaknesses, and describing their popular applications.
}
\input{Content/Table/table-1}

\subsection{Variational Autoencoders} 
\label{subsec:vae}
Variational Autoencoders (VAEs) represent a groundbreaking development in the field of generative models in machine learning \citep{kingma2013auto}. Fundamentally, a VAE is a probabilistic approach that combines aspects of neural networks with statistical modeling, allowing for efficient encoding and generation of complex data distributions \citep{doersch2016tutorial, kingma2013auto}. Unlike traditional autoencoders that merely compress and decompress data, VAEs introduce a probabilistic twist, enabling them to generate new data samples that are similar to the input data \citep{kingma2013auto}. This ability to model and sample from complex data distributions has found substantial applications in the realm of smart cities. Recent smart city studies have applied VAEs in multiple urban planning and environmental management applications, with the core capabilities of VAEs summarized as the following:
\begin{itemize}
  \item Missing Data Imputation: VAEs can facilitate the process of replacing missing data in a dataset with substituted values. The goal is to address gaps and poor quality in data to allow for effective analysis and decision-making. Many previous urban science applications employed the VAE to support data imputation and enrichment in the interest of improving the prediction and forecasting in traffic \citep{chen2021learning, boquet2020variational,pereira2020reviewing}, energy \citep{shen2021missing, liu2021two}, and urban water resources management \citep{zhang2023miss,wang2023edge,xie2019supervised}.
  \item Anomaly Detection: the process involves identifying unusual patterns or outliers in city-related datasets that deviate significantly from the norm. This process is crucial in urban contexts, as it aids in recognizing potential issues or abnormalities in various sectors. VAEs are utilized to improve the Detection of anomalies in complex urban data collected from IoT devices and smart infrastructure \citep{villegas2023iot,vu2020learning,mavikumbure2022anomaly}.
\end{itemize}
The versatility and effectiveness of VAEs in data enrichment make them an invaluable tool in advancing smart city initiatives and improving the availability and quality for developing data-driven solutions for sustainable and efficient urban living.

\subsection{\rev{Generative Pre-trained Transformers}} 
\label{subsec:transformer-Based-Models}
\rev{Generative pre-trained transformers (GPT) are a type of Large Language Models (LLMs)}, a groundbreaking innovation in natural language processing (NLP), and have revolutionized how machines understand and generate human language \citep{gillioz2020overview}. Characterized by their unique architecture, which relies on self-attention mechanisms, transformers facilitate the processing of data sequences, such as text, in a manner that captures contextual relationships more effectively than prior methods \citep{vaswani2017attention}. This architecture has been instrumental in the development of advanced language models like ChatGPT, which demonstrates exceptional capabilities in generating human-like text, answering questions, and even assisting in coding and analytics tasks \citep{feng2023investigating,roumeliotis2023chatgpt}.

In recent years, the scientific research community has shown a growing interest in transformer-based models, especially the \rev{OpenAI's ChatGPT and Google's BERT}, primarily due to their versatility and efficiency in understanding and contextualizing text for data and knowledge discovery, as well as their intelligent capability for generating codes in multiple programming languages \citep{sohail2023decoding}. In the urban science sector, GPT and its Application Programming Interface (API) have been used to support a variety of innovative applications in the following aspects:
\begin{itemize}
  \item Data Mining: analyzing and extracting text-data and image-data from social-media and crowed-sourcing platforms \citep{hu2023geo, kheiri2023sentimentgpt}.
  \item Autonomous Expert System: leveraging the \rev{GPTs'} capability to automate domain-specific data discovery \citep{li2023metaqa} and code generation \citep{li2023autonomous, feng2023investigating}, \rev{ontology and knowledge graph creation \citep{tupayachi2024towards}},
  and facilitate decision-support \citep{singh2023leveraging,tan2023promises} for complex environmental and urban problems.
\end{itemize}

\rev{Early versions of Large Language Models (LLMs) and their associated applications, such as Contextual Predictive Text systems (CPTs), faced significant limitations in their knowledge base due to the static nature of the data on which they were trained \citep{bubeck2023sparks}. Concerns are raised regarding the factuality of AI-Generated response to domain-specific questions that require the most current and dynamic information, and specilized knowledge \citep{cao2021hallucinated, bender2021dangers}. To address these shortcomings, there is a need to improve the quality and accuracy of the AI-generated content with vast factual knowledge retrieved from online knowledge base \citep{borgeaud2022improving}. Recent advancements in AI and computing technologies have introduced various tuning techniques for LLMs, together with the Retrieval-Augmented Generation (RAG) technique, to enhance the GPT' ability to access and utilize up-to-date data and specialized domain knowledge base \citep{lewis2020retrieval}, thereby improving their performance in completing tasks related to specialized domains and ensuring more updated and relevant responses with high factual accuracy \citep{wang2023survey}.}

\subsection{Generative Diffusion Models} 
\label{subsec:diffusion-models}
Diffusion Models, a class of generative models that have gained prominence in recent years, represent a novel approach in the field of machine learning for synthesizing high-quality data \citep{ho2020denoising,sohl2015deep}. 
These models operate on the principle of gradually transforming a distribution of random noise into a distribution of structured data by mimicking the process of diffusion in physics and have broken the long-time dominance of GANs \citep{yang2023diffusion}. The roots of Diffusion Models can be traced back to earlier works in Denoising Diffusion Probabilistic Models (DDPMs) \cite{ho2020denoising} and score-based generative models \citep{song2020score}. Still, the research in the late 2010s truly harnessed their potential in generating complex data such as images and audio.
In scientific research, diffusion Models hold significant promise due to their ability to generate realistic and varied data samples, improve likelihood estimation, and handle data with special structures \citep{yang2023diffusion,cao2022survey}. In the urban science and management sector, the recently emerging generative diffusion models are increasingly applied in smart applications to facilitate the simulation and modeling of urban dynamics and processes \citep{zhou2023towards,zhong2023guided}, as well as generative urban design and optimization \citep{du2023beyond,su2023floor,he2023generative}.

Generative diffusion models are also effective in data augmentation \citep{reutov2023generating} for generating smoothly varied versions of the original data and supporting anomaly detection and denoising. However, their applications in these areas are still rare in the urban science sector. The main strength of diffusion models is their ability to generate intricate, high-fidelity representations of complex systems, making them an invaluable tool for smart city development's predictive, autonomous planning and optimization. 

\subsection{\rev{Generative Model Comparison} }
\label{subsec:genAI-Comparision}
\rev{
Based on numerous literature reviews and surveys on popular GenAI models, we present a comparative summary of different generative AI models in Table \ref{tab:allGenAI}, highlighting their strengths, weaknesses, and applications. It is worth mentioning that many generic GenAI applications, such as the generation of images, vectors, and 3D models, can be tailored to generate specialized GIS data for smart city applications. This is because raster data, shapefiles, and 3D terrains are essentially spatially referenced images, vectors, and 3D models respectively. 
Theoretically, the currently popular applications of GenAI models for artistic and recreational purposes can be tailored to support urban science research and smart city applications. This can be achieved by training these models using urban datasets and simulation outputs.
}
\input{Content/Table/table-2}

%% file: Content/Table/table-1.tex
\begin{scriptsize}
\begin{table*}[hbt]
\footnotesize
\caption{\rev{Comparison between GAN and its variants.}} \label{tab:gans} 
\begin{tabular*}{\textwidth}{|p{1.4cm}|p{3.8cm}|p{3cm}| p{3cm}| p{3.05cm}| }
 \hline
 \multicolumn{1}{c}{\cellcolor{black!10}\textbf{Model}}  & \multicolumn{1}{c}{\cellcolor{black!10}\textbf{Rationale}} &  \multicolumn{1}{c}{\cellcolor{black!10}\textbf{Strengths}} &  \multicolumn{1}{c}{\cellcolor{black!10}\textbf{Weaknesses}}  &  \multicolumn{1}{c}{\cellcolor{black!10}\textbf{Popular Applications}} \\
\hline
 GANs & Two neural networks (Generator \& Discriminator) compete against each other. & Produce high-quality, realistic samples \citep{shmelkov2018good}. & Mode collapse and training instability \citep{thanh2020catastrophic}. 
 &  Image, vector, and raster data generation, style transfer, data augmentation \citep{wu2022generative}. \\ 
 \hline
 cGANs & GANs with additional information (labels) provided to both generator and discriminator. & Generate class-specific samples, improved control over output \citep{rangwani2022improving}. & Requires labeled data, potential for overfitting \citep{gauthier2014conditional}. &  Synthetic and conditional data generation, simulation scenario generation \citep{durgadevi2021generative}. \\
 \hline  
 CycleGANs & Use two GANs to learn mappings between two domains without paired examples. & Generate high-quality sample and capture dependencies between features in the data and images \citep{ma2024comparison}. & Training complexity and potential for artifacts \citep{ma2024comparison}. & Aerial-to-map translations, style transfer and unsupervised image-to-image translation \citep{ma2024comparison,fontanini2022avoiding}. \\
 \hline
\end{tabular*}
\end{table*}
\end{scriptsize}

%% file: Content/Table/table-2.tex
\begin{scriptsize}
\begin{table*}[hbt]
\footnotesize
\centering
\caption{\rev{Comparative summary of the strength, weakness, and applications of popular GenAI models.}} \label{tab:allGenAI} 
\begin{tabular*}{\textwidth}{|p{1.1cm}|p{2.5cm}|p{4.5cm}| p{3cm}| p{3.15cm}| }
 \hline
\multicolumn{1}{c}{\cellcolor{black!10}\textbf{Model}}  & \multicolumn{1}{c}{\cellcolor{black!10}\textbf{Rationale}} & \multicolumn{1}{c}{\cellcolor{black!10}\textbf{Strengths}} & \multicolumn{1}{c}{\cellcolor{black!10}\textbf{Weaknesses}}  & \multicolumn{1}{c}{\cellcolor{black!10}\textbf{Popular Application}} \\
\hline
 GANs & Two neural networks (Generator and Discriminator) compete against each other & Produce high-quality and realistic samples \citep{shmelkov2018good}. & Mode collapse and training instability \citep{thanh2020catastrophic}. &  Image, vector, and raster data generation, style transfer, data augmentation \citep{wu2022generative}. \\ 
 \hline
 VAEs & Latent space modeling via encoder-decoder architecture with variational inference. & Robust training through explicit modeling of data distribution, the usage of stochastic gradient descent, and regularization through Kullback-Leibler (KL) divergence \citep{cemgil2020autoencoding, kingma2017variational}.  
 & Blurriness in generated images, less sharp results \citep{zheng2020conditional}. & Data compression, image generation, anomaly detection, and dimension reduction \citep{mak2023application}. \\
 \hline
 GPT & Attention mechanisms to capture dependencies across input sequences. & Natural language understanding and generation, contextual understanding, and few-shot and zero-shot learning with minimum training data \citep{ray2023chatgpt}. 
 & Limited sequential modeling, high computational cost, and potential bias \citep{kalyan2023survey, brown2020language}. & Natural language processing, image generation, human language understanding, code generation, translation tasks \citep{kalyan2023survey}  \\
 \hline
 Diffusion Models & Gradually unfold the data distribution through iterative refinement. & Generate high-quality samples that capture complex dependencies and may outperform GANs \citep{dhariwal2021diffusion}. & Computationally intensive and sensitive to initialization \citep{song2020score}. & Generation of image, audio, 3D model, and graph-structured data generation, as well as image inpainting and anomaly detection \citep{song2020score}. \\
 \hline
\end{tabular*}
\end{table*}
\end{scriptsize}

%% file: Content/30-applications.tex
\section{GenAI in Smart City Applications} \label{sec:Dataset}
This section will provide a structured review of how recent advancements in GenAI are revolutionizing key areas of urban science research and smart city applications. These areas include traffic and transportation management, mobility research, building operations, energy management, and infrastructure planning. We will primarily focus on the applications of various GenAI models, such as GANs, VAEs, \rev{GPTs}, and generative diffusion models. \rev{Our aim is to address research question Q3 by exploring how GenAI technologies are applied in the smart city sector through comprehensive reviews of previous studies.}

\begin{figure*}[htb]
 \centering
\includegraphics[width=\textwidth]{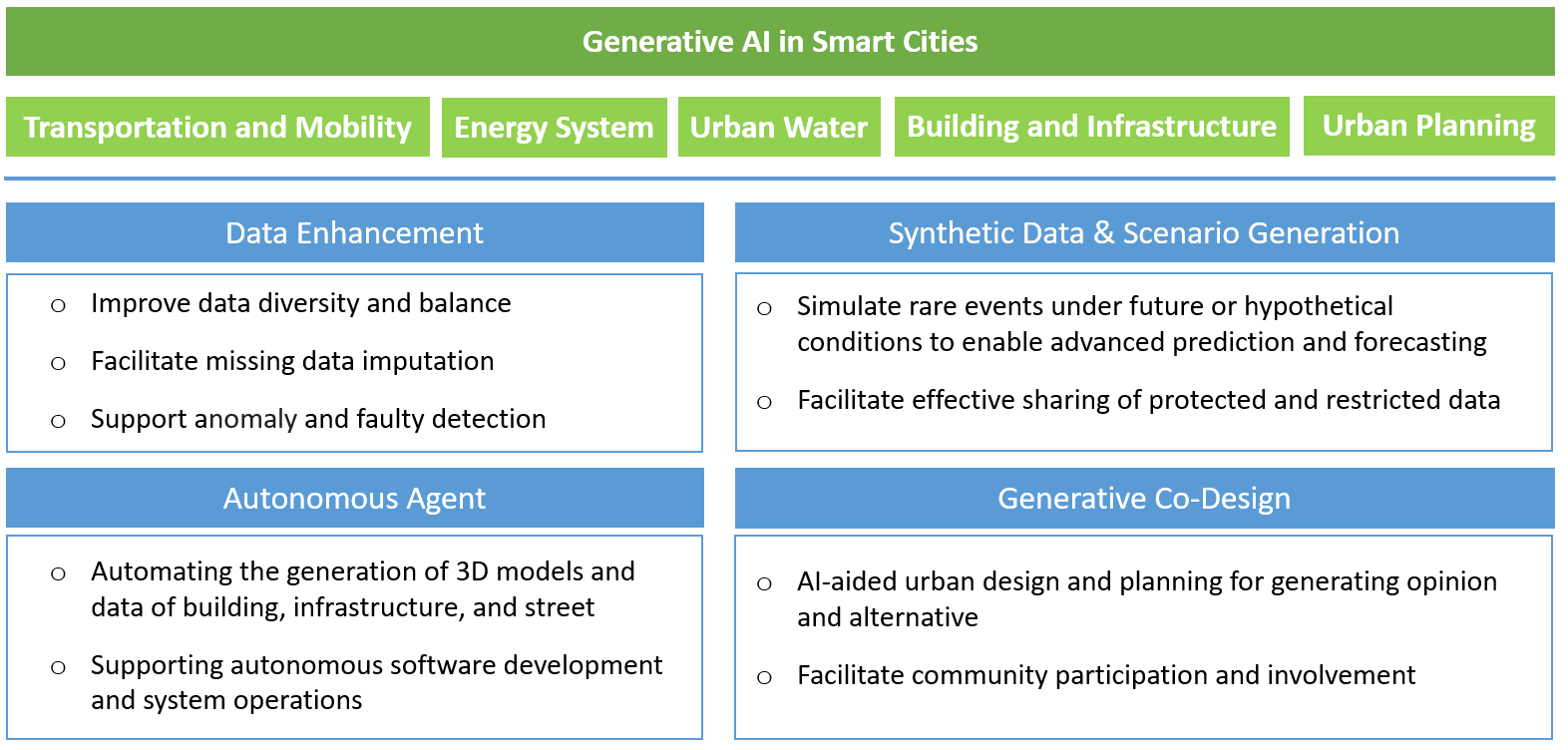}
 \caption{GenAI application areas in major smart city sectors.}
 \label{fig:overview}
\end{figure*}

The following subsections will delve into specific GenAI applications in each of the aforementioned urban application areas (as depicted in Figure \ref{fig:overview} and defined in \ref{tab:udt}).  \rev{These subsections aim to highlight GenAI's ability to address challenges in urban digital twin applications discussed in Section \ref{sec:udt-challenges}. Within each subsection, we will select a handful of applications to discuss in detail, focusing on their methodology and implementation strategy. This selection is based on the study's potential for integration into urban digital twins, with a preference for data-driven applications that align with the digital twin's primary functions of enabling situational awareness, predictive analysis, and informed decision support. }

\subsection{Mobility and Transportation Management}
\rev{\rev{From the 259 articles selected for their relevance to urban transportation, we performed an additional screening of their titles, keywords, and abstracts. This allowed us to identify the primary GenAI use cases that can be integrated into urban digital twins within the context of intelligent transportation systems and smart mobility. The various use cases are detailed in the following subsections.}}

\subsubsection{Transportation Data Augmentation} 
\label{subsec:trans-data-augmentation}
Data augmentation refers to techniques used to increase the amount of data by adding slightly modified copies of already existing data or newly created synthetic data from existing data \citep{lemley2017smart}. This is particularly common in machine learning and deep learning, where more training data leads to better model performance. Data augmentation is primarily about creating additional data points to improve the robustness and accuracy of models in tasks like image recognition, classification, and predictions \citep{moreno2020improving}. GANs and VAEs are prevalent in smart city applications for data augmentation, given their proficiency in generating high-quality synthetic data. However, generative diffusion models, known for their strengths in denoising and data augmentation, are primarily utilized in medical research rather than urban science. This suggests a potential untapped area for these models in urban applications, where their unique capabilities could offer novel solutions to complex urban challenges. \rev{Many previous studies have applied GenAI models to augment transportation data.} \citet{zhou2021improving} introduces the Trajectory Generative Adversarial Network (TGAN) to address challenges in identifying and linking human mobility patterns, particularly in location-based social networks (LBSNs). By combining adversarial learning and spatio-temporal data analysis, TGAN significantly enhances trajectory-user linking (TUL) by generating synthetic trajectories that closely mimic real human movement patterns. This approach effectively overcomes issues of data sparsity and imbalanced label distribution in trajectory data, demonstrating superior accuracy in TUL compared to existing methods. The use of GANs for data augmentation in this context marks a substantial advancement in human mobility analysis, offering promising implications for various applications, including urban planning and personalized services. \citet{jilani2022traffic} presents an innovative approach to traffic congestion analysis using GANs for data augmentation. The research introduces a five-layer convolutional neural network model that leverages GAN-enhanced datasets for improved accuracy in classifying traffic congestion. This method significantly improves identifying congestion patterns, achieving a 98.63\% accuracy rate. The study highlights the efficacy of GANs in generating realistic synthetic images, thereby enriching the dataset for more effective traffic management and planning applications. \citet{dabboussi2023gan} explores an innovative approach using GANs for data augmentation in traffic management. The study presents a novel GAN-based model that successfully generates synthetic traffic data, addressing the critical issue of data scarcity in Intelligent Transportation Systems (ITS). This method demonstrates a significant improvement in traffic prediction models, enhancing their performance through the addition of realistic, GAN-generated data. The research contributes to developing more intelligent and sustainable transportation systems, underlining the potential of GANs in transforming ITS applications. \citet{islam2021sensor} presents a novel data augmentation technique using a VAE for improving transportation mode classification with limited training data. The VAE's decoder is adept at generating synthetic data that closely resembles real smartphone sensor data, effectively enlarging the dataset for machine learning models. This method enhances classification accuracy and addresses the imbalance in datasets by boosting underrepresented classes. Its application in activity recognition using smartphone sensors holds significant potential for vehicle-to-pedestrian (V2P) systems and smart mobility solutions. \citet{islam2021applying} presents novel deep learning applications in urban transportation systems. It showcases the use of VAEs and GANs for augmenting data in scenarios with limited datasets. The study highlights the capability of these deep learning models to generate synthetic data, which closely resembles real-world data, thus enhancing model training and predictive performance in transportation applications like crash data analysis and traffic pattern prediction. \citet{islam2021crash} introduces a VAE-based data augmentation technique to balance crash and non-crash event datasets in traffic systems. Due to the extreme imbalance in these datasets, traditional machine learning models tend to underperform. The VAE encodes events into a latent space, enabling the generation of synthetic crash data that statistically mirrors real data. This method outperforms traditional oversampling techniques like SMOTE and ADASYN, improving specificity and sensitivity in crash prediction models. It also addresses overfitting issues, showcasing VAE's capability in generating precise, balanced datasets for effective machine learning applications in traffic safety.

\subsubsection{Missing Data Imputation} 
\rev{Missing data imputation using GenAI involves utilizing advanced generative models, such as VAEs or GANs, to predict and fill in missing values within a dataset \citep{shahbazian2023generative}. These models learn the underlying distribution of the complete data and generate plausible values for the missing entries, thereby enhancing the overall quality and integrity of the data \citep{shammasi2024enhancing}. By capturing complex patterns and relationships in the data, GenAI-based imputation provides more accurate and reliable estimates compared to traditional methods. Among previous studies, GenAI models have been widely used in intelligent transportation systems and smart mobility systems to facilitate the imputation of traffic and mobility data. These models effectively address missing data issues by generating plausible data points, thereby enhancing the accuracy and reliability of data analysis and decision-making processes in urban transportation networks. \citep{chan2023missing}. \citet{boquet2020variational} introduces a unified model leveraging VAEs to address key challenges in traffic forecasting. The study highlights the use of VAEs for generative modeling, enabling the imputation of missing traffic data, effective dimension reduction, model selection, and anomaly detection. By learning the latent space representation of traffic data, the VAE improves the accuracy and reliability of predictions, compresses data for computational efficiency, and identifies meaningful patterns and anomalies. Extensive experiments on real-world traffic datasets validate the model's superior performance over traditional methods, demonstrating its potential to enhance intelligent transportation systems. \citet{zhang2024spatial} introduces a novel approach utilizing GenAI models to address the challenge of missing traffic data. The study proposes a dynamic multi-level generative adversarial network (MLGAN) model that captures the spatial-temporal dependencies of traffic data through a combination of bidirectional gated recurrent units (BGRU) and graph convolutional networks (GCN). By leveraging these advanced generative models, the MLGAN effectively imputes missing traffic data, even under conditions of high fluctuation and non-uniform distribution of missing values. The research demonstrates that the MLGAN model outperforms traditional imputation methods, providing more accurate and stable imputation results, thereby significantly enhancing the reliability and efficiency of urban traffic management systems. Extensive experiments using real-world data from the PeMS dataset validate the superior performance of the proposed model compared to existing approaches. Many similar GenAI-powered applications used varients of GAN to impute traffic data \citep{yuan2022stgan, kazemi2021igani, yang2021st, huang2023deep, chen2023prediction}. A few data have applied GenAI models to imputing missing data in urban mobility systems, such as bike-sharing data \citep{xiao2021efficient} and  public transit data \citep{kim2022imputing}.
}

\subsubsection{Generation of Synthetic Mobility Data }
\rev{In human mobility studies, generative models are widely used to generate synthetic trajectories that can realistically reproduce mobility patterns and enable predictions of future mobility trends. Synthetic data generation involves creating entirely new data points that mimic the statistical properties and structure of the original dataset \citep{offenhuber2024shapes}. GenAI models, such as GANs and VAEs, are often used for this purpose. The primary goal is to expand the dataset, providing a larger volume of high-quality data for training machine learning models. This approach is particularly useful when real data is scarce, expensive to collect, or poses privacy concerns. In comparison, the generative task focuses on creating entirely new data points, while data augmentation modifies existing data to produce variations, and data imputation fills in gaps within an existing dataset. Each technique addresses different challenges in data preprocessing and enhancement, contributing to the overall improvement of data quality and utility.
}

\rev{As examples}, \cite{luca2021survey} have provided a comprehensive survey on the GenAI applications for creating synthetic trajectories. As examples, \citet{smolyak2020coupled} uses GANs for anomaly detection in human mobility data. The authors introduce the Infinite Gaussian Mixture Model (IGMM) coupled with bidirectional GANs (BiGANs), an approach that generates realistic synthetic human mobility data and effectively detects multimodal anomalies. This method marks a significant advancement in anomaly detection, addressing the challenges of multimodal data patterns in human mobility and improving the accuracy of identifying anomalous behaviors or trajectories.
\citep{kosaraju2019social} presents a novel approach to pedestrian trajectory forecasting. It integrates a graph-based GAN with a graph attention network (GAT) and Bicycle-GAN to generate realistic multimodal trajectory predictions as future scenarios. This method effectively models the social interactions of pedestrians and accounts for the multimodal nature of human movement. The research demonstrates significant advancements in trajectory forecasting, contributing to applications like autonomous vehicle navigation and urban planning. 
\citep{huang2019variational} utilizes a Sequential Variational Autoencoder (SVAE) for reconstructing human mobility trajectories. This approach is novel in its integration of a Variational Autoencoder (VAE) with a sequence-to-sequence model, enabling the efficient capture of salient features in human mobility data. The SVAE model demonstrates the capability of VAEs in approximating complex, high-dimensional distributions like urban mobility patterns. This innovation is crucial for enhancing trajectory reconstruction in urban planning and intelligent transportation systems, addressing challenges of data sparsity and privacy concerns. \citet{bao2020covid} introduces an innovative approach using Conditional GANs to estimate human mobility during the COVID-19 pandemic. This research significantly contributes by addressing the challenge of estimating human mobility in rapidly changing social contexts with limited training data. The study employs a domain-constraint correction layer within the GAN's generator, enhancing the learning process and improving the quality of generated mobility predictions. This method demonstrates the potential of GANs in generating reliable spatiotemporal mobility data under uncertain and dynamically changing conditions, a crucial advancement in predictive modeling during public health crises. \citet{wang2020seqst} introduces SeqST-GAN, a model utilizing Sequence-to-Sequence (Seq2Seq) Generative Adversarial Networks for predicting urban crowd flow. It uniquely addresses the complex spatial-temporal correlations in urban environments by treating crowd flow data as sequences, akin to frames in a video. The combination of Seq2Seq and GANs enables the model to generate accurate multi-step predictions, overcoming the challenge of blurry predictions often seen in deep learning models. This innovative use of GANs demonstrates their capability in generating sequential, high-dimensional data, marking a significant advancement in urban planning and traffic management.

\subsubsection{\rev{Traffic Simulation Scenario Generation}}
In many previous efforts, generative models have been developed to create synthetic data for planning and simulation scenarios. Numerous studies have specifically applied variants of GANs for these purposes. In terms of simulating traffic flow and dynamics, \citep{wu2020spatiotemporal} explores a novel approach for generating realistic traffic flow scenarios using GenAI. The study leverages a hybrid model combining Long Short-Term Memory (LSTM) networks with GANs, termed LSTM-GAN. This model captures the temporal dynamics and spatial dependencies of traffic data, enabling the creation of spatiotemporal traffic scenarios for multiple horizons and locations. The GenAI component, specifically the GAN, plays a crucial role in learning from historical traffic data to generate new, realistic traffic flow scenarios. These synthetic scenarios can aid in traffic infrastructure design, road planning, and virtual training environments for intelligent driving systems. The research demonstrates that the LSTM-GAN model outperforms traditional GAN models by better capturing the temporal and spatial correlations inherent in traffic data, thereby providing more accurate and useful synthetic data for intelligent transportation systems. \citet{mladenov2022adversarial} explores a novel application of GANs in traffic simulation. The study introduces an adversarial variational inference technique for calibrating Origin-Destination (OD) matrices in urban traffic simulators. This method is significant for its ability to mitigate the underdetermination problem in traffic simulation, leveraging GANs to match higher-order moments of field data. This innovative approach enhances the accuracy and robustness of traffic simulations, which is essential for effective urban traffic management and planning. 

\rev{There are also a few similar studies that use GAN-based models to generate traffic simulation scenarios to support autonomous driving research. \citet{demetriou2023deep} utilizes GANs to create realistic driving scenarios for autonomous vehicle testing. This research demonstrates the innovative use of GANs in generating complex and diverse driving conditions, aiding in the development and validation of autonomous driving systems. The study highlights GAN's capability to simulate various road environments and traffic situations, which is crucial for ensuring the safety and reliability of autonomous vehicles in real-world conditions. \citet{li2021scegene} introduces a novel approach for generating realistic and diverse traffic scenarios using GenAI. The study presents SceGene, a dynamic traffic scenario generation algorithm inspired by biological intelligence, specifically genetic inheritance and mutation processes. SceGene applies biological processes like crossover, mutation, and natural selection to create new traffic scenarios. This method involves encoding traffic scenario elements into a genotype-like representation, generating new scenarios through genetic operations, and repairing abnormal scenarios using a microscopic driving model. By leveraging these GenAI techniques, SceGene can produce highly realistic and diverse traffic scenarios in an interpretable and controllable way, significantly improving the efficiency and coverage of simulation-based autonomous driving tests. The generated scenarios enhance the robustness and safety of autonomous driving algorithms by providing a comprehensive range of testing conditions. Apart from GAN-based applications, \citet{kempinska2019modelling} explores the use of deep learning, specifically VAEs, in modeling urban street networks. It demonstrates how VAEs can effectively capture key urban network metrics in low-dimensional vectors and generate complex urban forms. This approach, capitalizing on the deep learning capability of generating realistic images, marks significant progress in data-driven urban mobility modeling, leveraging the increasing availability of street network data to enhance the quantitative analysis of urban patterns. } 

\rev{With the recent advancement in LLMs, a few emerging studies explores the use of GPT to generate scenarios. As a exmaple, \citet{guzay2023generative} explores the innovative use of Large Language Models (LLMs) like GPT-4 to generate traffic scenarios for simulation purposes. The study highlights how LLMs, typically used in natural language processing, can be adapted to create XML-based simulation files for tools like SUMO (Simulation of Urban Mobility). By leveraging the generative capabilities of LLMs, the researchers aim to simplify the scenario creation process, making it more fluent and flexible compared to traditional methods. The approach allows users to describe traffic scenarios linguistically, which are then processed by the model to produce accurate simulation files. This application of GenAI not only streamlines the development of traffic scenarios for safety testing, autonomous vehicle training, and regulatory compliance but also addresses challenges related to the complexity and rigidity of existing simulation setup methods.}

\rev{ Apart from the four major types of GenAI models, our literature search has also identified the application of other types of GenAI models. \citet{tan2021scenegen} presents a novel approach to generating realistic traffic scenarios using GenAI. The study introduces SceneGen, a neural autoregressive model that generates traffic scenes by sequentially inserting actors such as vehicles, pedestrians, and bicyclists into a given environment based on the ego-vehicle's state and a high-definition map. This method avoids the need for hand-crafted rules and heuristics, allowing for more complex and diverse traffic scenarios that better mimic real-world conditions. The GenAI model used in SceneGen captures the spatial and temporal dependencies of traffic scenes, significantly improving the fidelity of the simulations. The results demonstrate that SceneGen can generate realistic traffic scenarios that enhance the training and testing of self-driving vehicles, providing more accurate and scalable solutions compared to traditional methods.}

\subsection{Urban Energy Systems} 
\rev{Among the 479 selected articles refined for their relevance to urban energy systems, we conducted a further screening based on their titles and abstracts to summarize the major GenAI use cases that can be incorporated into urban digital twins in the context of urban energy management and grid operation. These use cases are discussed in the following subsections.}

\subsubsection{\rev{Energy Data Enhancement}}
\rev{Data enhancement encompasses various techniques aimed at improving the quality, quantity, and diversity of data. Among the selected studies, two frequently used GenAI-powered techniques are data augmentation and missing data imputation. While some autoencoder-based GenAI models, such as VAEs, are capable of performing other data enhancement tasks, such as dimensionality reduction and anomaly detection, these features do not leverage the generative capabilities of the AI. Consequently, studies related to these features are not discussed in this section.}

\rev{Data augmentation techniques, powered by GenAI, are becoming increasingly pivotal in building energy modeling and power grid applications. These innovative methods have substantially improved the accuracy of predictions and forecasts, thereby optimizing the management of urban energy systems.} \citet{wu2021gan} presents a novel approach to forecasting energy consumption using a GAN-enhanced ensemble model. This method combines GANs with ensemble learning, demonstrating significant improvements in forecasting accuracy. The study explores the use of GANs for data augmentation, which effectively enriches the training dataset with high-quality, diverse samples. This approach enhances the model's robustness and generalization capabilities, showcasing the utility of GANs in improving predictive models in energy management systems. The \citet{boicea2022novel} explores using GANs for augmenting data in power load forecasting. It demonstrates GAN's capability to expand a dataset, enhancing the predictive accuracy of a Convolutional Neural Network model used for load forecasting. This approach addresses the challenge of limited sample size in power load forecasting, showing GANs' effectiveness in data augmentation and their potential in improving forecasting accuracy in energy management systems. \citet{fan2022novel} utilizes deep learning, specifically deep generative modeling, for data augmentation in building energy prediction. This approach significantly improves the accuracy of short-term energy predictions in buildings. The study leverages the capabilities of deep learning to generate synthetic but realistic energy data, addressing the challenge of limited high-quality training data. This innovation demonstrates the potential of deep learning in enhancing data-driven models for smart building operation management, especially in scenarios with data scarcity.

\rev{A few studies have utilize GenAI models to facilitate missing data imputation. \citet{hwang2024cc} proposes a novel method for addressing missing data in electricity consumption datasets, crucial for effective energy management in smart grid environments. The study introduces the CC-GAIN model, which integrates clustering, classification, and GANs to impute missing data. This model leverages time-series and pattern features of electricity consumption data to enhance imputation accuracy. The CC-GAIN model significantly outperforms existing imputation techniques by effectively handling diverse types and rates of missing data, making it a robust solution for maintaining high-quality data essential for smart grid operations. The research highlights the model's superior performance through extensive evaluations and ablation studies, demonstrating its potential to improve data reliability and support sustainable energy management.
\citet{ryu2020denoising} explores a deep learning approach for addressing missing data in smart meter datasets. The study introduces a framework using a Denoising Autoencoder (DAE) to impute missing electricity consumption values. This GenAI technique is compared with traditional imputation methods like linear interpolation and historical averaging, as well as other generative models such as the VAE and Wasserstein Autoencoder (WAE). The DAE-based approach significantly outperforms these methods, showing up to 28.9\% lower point-wise errors and up to 56\% lower daily-accumulated errors. The paper highlights the robustness and accuracy of the DAE model in reconstructing incomplete daily load profiles, thus enhancing the reliability of smart grid data analytics.
}

\subsubsection{Data Synthesis and Scenario Generation}
Synthetic data supports the planning, management, and operation of urban energy systems by providing realistic, diverse datasets for modeling and simulating various energy scenarios\citep{li2020building, roth2020syncity}. These datasets can facilitate the optimization of energy distribution, demand prediction, and efficient resource allocation, reducing carbon-emission in cities and improving efficiency in grid systems.

As detailed examples, \citet{wang2020generating} introduces an innovative approach to generate realistic building electrical load profiles. Utilizing GAN, the research addresses the challenge of generating load profiles that accurately reflect dynamic and stochastic behaviors of real buildings. The study showcases GAN's capability to anonymize smart meter data, maintaining important statistical information while ensuring privacy. This advancement offers significant implications for smart meter data analytics, enhancing the integration of buildings with the electric grid and promoting sustainable energy management.
\citet{dong2022data} discusses a novel data-driven method for generating renewable energy production scenarios using controllable GAN. It focuses on addressing the challenges in power system planning and operation due to uncertainties in wind and solar energy. The proposed GAN model learns the stochastic and dynamic characteristics of renewable resources, enabling the creation of controllable, interpretable scenarios. This method enhances renewable energy modeling, offering significant contributions to the efficiency and reliability of power systems with high renewable penetration.
\citet{pan2019data} introduces an innovative application of deep learning in the field of electric vehicle (EV) load profiles generation. It employs a Variational Auto-Encoder (VAE), a deep learning model, to create EV charging load profiles. This approach is significant because it moves away from traditional methods that heavily rely on probabilistic models and manual data sampling. The VAE model, by learning the inherent characteristics of EV load data, generates diverse load profiles that accurately reflect temporal correlations and probability distribution of actual loads. This methodology signifies a noteworthy leap in using deep learning for more efficient and accurate energy management in the context of increasing EV usage.
\citet{jiang2018scenario} presents a novel approach for wind power scenario generation using GANs. This research stands out for its improvement of GANs, particularly in enforcing Lipschitz continuity through gradient penalty and consistency terms. These enhancements address common issues like overfitting and training instability, leading to more efficient and stable model training. The study demonstrates the effectiveness of this approach in generating high-quality scenarios for wind power, offering significant potential for renewable energy modeling and decision-making under uncertainty.
\citet{chen2018model} introduces an innovative approach for renewable energy scenario generation using GANs. This study highlights the ability of GANs to create realistic, data-driven scenarios, capturing the spatial and temporal correlations of renewable power plants. The method is particularly notable for its efficiency and scalability, offering a significant advancement in renewable energy modeling, especially for wind and solar power generation. This work demonstrates the potential of GANs in effectively handling the complexities and variabilities in renewable energy data.

\subsubsection{\rev{Grid Operations}}
\rev{One study has explored the capability of large language models (LLMs), especially GPTs, in developing AI agents to facilitate decision support and the operation of power grids and energy facilities. \citep{eGridGPT2024} explores the integration of large language models (LLMs) in power grid control rooms to aid decision-making processes. eGridGPT is designed to support grid operators by analyzing data, suggesting actions, and simulating scenarios to enhance grid reliability amidst increasing adoption of renewable energy and distributed energy resources. The report highlights eGridGPT’s potential to manage operational challenges posed by inverter-based resources and external threats like extreme weather and cyber-attacks. Emphasizing the need for trustworthiness, the system incorporates explainable AI principles, digital twin simulations, and a human-in-the-loop framework to ensure safe, transparent, and reliable operations. The report also discusses the importance of accurate data, ethical AI usage, and low-budget solutions for smaller utilities, concluding that eGridGPT can significantly contribute to a resilient and affordable future energy system.}

\rev{One study, \citet{lin2023knowledge}, discusses the application of GenAI models in constructing and utilizing knowledge graphs for power grid equipment management. The study employs GANs to generate realistic data that populates the knowledge graph, ensuring comprehensive coverage of various equipment and scenarios. This generative approach allows for the synthesis of high-quality, detailed data that enhances the accuracy and reliability of the knowledge graph. The enhanced knowledge graph supports better decision-making and predictive maintenance in power grid operations, demonstrating the significant impact of GenAI technologies in managing and optimizing power equipment.}

\subsection{Urban Water Management} 
\rev{Among the selected articles, only 105 studies are related to water resource management topic. Through a thorough manual screening process, which removed irrelevant studies that focuses on agriculture watershed management, environment remediation, chemical hydrology, and fluid dynamics, we discovered that GenAI applications in urban water management, which can be integrated into urban digital twins, are relatively limited, especially when compared to its broader use in other urban sectors.}
\subsubsection{Sewer and Drainage Management}
\rev{Sewer is considered as part of urban drainage system, it involves the collection, transportation, treatment, and disposal of wastewater and sewage from residential, commercial, and industrial sources. Effective sewer management is crucial for maintaining public health, protecting the environment, and ensuring the overall sustainability of the urban water system.}
\citet{koochaligenerative} explores the innovative usages of GANs in urban water management, specifically in improving combined sewer overflow predictions, such as depth and flow rate. GANs are used for data augmentation, generating synthetic time series to balance datasets, thus enhancing the accuracy of prediction models. The study demonstrates that models trained with synthetic data generated by GANs can perform comparably to those trained with real data. The study proposed novel deep-learning approaches that can address challenges like data scarcity in urban water management. \citet{situ2021automated} presents a novel approach for sewer defects detection. It utilizes state-of-the-art Style-based Generative Adversarial Networks (StyleGANs) to generate high-quality synthetic images of sewer defects. These images are then used to train and fine-tune seven well-known CNN models for automated sewer defects detection. The study demonstrates that both variants of StyleGANs are efficient in producing diverse and detailed images, significantly enhancing the performance of deep learning models in sewer defects detection. 

\subsubsection{\rev{Water Distribution System}}
\rev{
An urban water distribution system is responsible for transporting and delivering potable water from treatment facilities to residential, commercial, and industrial users throughout the city. It ensures a reliable supply of clean water through an extensive network of pipes, pumps, and storage facilities, maintaining adequate pressure and quality standards to meet the daily needs of the urban population.
\citet{mcmillan2024domain} presents a novel framework utilizing GenAI models to enhance leakage detection in water distribution systems. The study employs a domain-informed VAE to reduce the dimensionality of water flow time series data into a two-dimensional latent variable space. This allows the model to capture distinct characteristics of leakage versus non-leakage flow. The VAE is combined with a Support Vector Machine (SVM) classifier to accurately distinguish between leakage and regular flow, leveraging the generative capabilities of the VAE to create realistic data representations. By using this GenAI approach, the framework achieves over 98\% accuracy in classifying leakage events, significantly improving the reliability and efficiency of leakage detection. This method supports the development of self-healing water infrastructure systems by enabling real-time, automated detection and response to leaks, thus enhancing sustainability and resilience in urban water supply management. \citet{li2023generative} explores the use of GANs to enhance the detection of contamination events in water distribution networks. The study leverages GANs to model the spatial and temporal correlations of water quality data collected from multiple sensors across different locations. By training a generator and a discriminator, the GAN-based method produces anomaly scores that identify deviations from normal water quality patterns, indicating potential contamination. The model integrates these scores using Bayesian sequential analysis to update the probability of contamination events in real-time. This approach demonstrates significant improvements in detection accuracy and robustness compared to traditional methods, showcasing the potential of GenAI in maintaining safe and reliable water supplies.
\citet{fu2022role} focuses on the application of GANs for tasks like predicting water system behaviors, improving operational efficiency, and enhancing decision-making processes. The paper underlines the effectiveness of GANs in handling complex, data-rich scenarios typical in urban water systems, demonstrating how these advanced AI models can contribute to more sustainable and effective water management strategies.}

\subsection{Building and Infrastructure Management}
\rev{From the 81 articles selected for their relevance to building and urban infrastructure management, we performed an additional screening based on their titles and abstracts. This process allowed us to identify the primary GenAI use cases that can be integrated into urban digital twins, particularly in the context of the urban built environment, BIM, CIM, and 3D city modeling. The various use cases are detailed in the following subsections.}
 
\subsubsection{\rev{Asset Operation and Maintenance}}
\rev{
The operation and maintenance of buildings and infrastructure have been revolutionized by 5G networks and IoT techniques, which generate large volumes and a variety of data characterizing the operating conditions of assets within the built environment. In this sector, GenAI models play a crucial role in augmenting and enriching building and infrastructure data to improve situational awareness, enhance simulation capabilities, and inform decision-making for asset management. By generating data that replicates a vast array of environmental and usage conditions, these AI models empower the development of robust solutions, including innovative designs and optimization strategies. Their applications span a broad spectrum, encompassing energy and equipment management, structural detection, and maintenance forecasting, leading to more resilient and efficient operations and maintenance of buildings and urban infrastructure.}

\rev{As examples,} \citet{abdollahi2020building} presents a groundbreaking approach in urban planning by integrating GANs, Genetic Optimization Algorithms (GOAs), and Geographic Information Systems (GIS). This study underscores the capability of deep learning, particularly GANs, in generating complex, high-dimensional urban building images. The innovative use of GANs for data augmentation, combined with the optimization strength of GOAs and the spatial analysis of GIS, demonstrates a novel framework in urban planning, enhancing efficiency, scalability, and decision-making in urban development.
\citet{li2023machine} addresses a significant challenge in the Structural Health Monitoring (SHM) of civil infrastructure using machine learning (ML), where a lack of comprehensive datasets for training models under various structural damage conditions is a common issue. To overcome this, the research introduces a novel data augmentation strategy within a vibration-based SHM framework, employing a Conditional Variational Autoencoder (CVAE). The CVAE is used to generate additional data samples of power cepstral coefficients of structural acceleration responses in diverse damage scenarios. This is achieved by training the CVAE to model different statistical distributions of these coefficients, using newly defined conditional variables. Notably, the decoder of the CVAE, once trained, is then used independently for the generation of new samples, enhancing the data augmentation process.
\citet{hoeiness2021positional} utilizes augmented GANs for predicting wind flow in urban environments. This innovative approach rephrases the problem of 3D flow fields into a 2D image-to-image translation challenge. The study explores the use of GANs, particularly Pix2Pix and CycleGAN, along with U-Net autoencoders, demonstrating GAN's ability to model complex spatial relationships and generate accurate wind flow predictions. This research represents a significant advancement in urban planning and pedestrian comfort analysis for improving street and infrastructure design. \rev{\citet{kastner2023gan} presents an innovative GAN model for predicting urban wind flow in built environment. This GAN, designed for image-to-image translation, efficiently maps input urban geometry (e.g, building and infrastructure) to predicted wind flow patterns. By utilizing Pix2Pix GAN, integrated with a voxelization process and distance-to-object encoding, the model offers rapid and accurate predictions of urban wind flow. This approach significantly reduces the computational cost and time associated with traditional Computational Fluid Dynamics (CFD) simulations, making it a promising tool for urban design and microclimate analysis.}

\rev{In addition to data enhancement, recent studies in the building technology sector have begun exploring the capability of large language models (LLMs), specifically GPTs, to facilitate the construction, operation, and maintenance of buildings and infrastructure. These models offer advanced analytical and predictive capabilities, contributing to more automated, efficient, and effective management practices in the built environment. \citet{ahn2023alternative} investigates the application of a pre-trained large language model, specifically ChatGPT, to control HVAC systems in buildings to enhance energy efficiency and maintain indoor air quality. Traditional HVAC control methods often rely on rule-based or model-based approaches, which can be limited in flexibility and efficiency. By leveraging the capabilities of GenAI models such as ChatGPT, this study explores an innovative method for optimizing HVAC operations. ChatGPT, trained with reinforcement learning from human feedback and fine-tuned with proximal policy optimization, is utilized to make sequential decisions for HVAC system operations. The study compares ChatGPT-based control with a deep Q-network (DQN) model-free control and a baseline operation in an EnergyPlus simulation of a reference office building. Results show that ChatGPT's approach, which cycles through adjusting set-point temperatures and damper openings, can achieve substantial energy savings while maintaining CO2 concentrations below 1000 ppm. Although DQN control slightly outperforms ChatGPT in terms of energy savings (24.1\% vs. 16.8\%), ChatGPT's performance without additional training demonstrates its potential for high generalization and practical application in real-world settings. This study highlights the promise of GenAI models in enhancing building management systems and contributing to more sustainable urban environments. 
\citet{zhang2024automated} explores the application of GenAI, specifically GANs and VAEs, to optimize building energy systems, particularly HVAC operations. By generating realistic synthetic datasets that mimic actual operational data, the study aims to enhance energy efficiency, enable predictive maintenance, and preserve data privacy. The integration of these generative models into building management systems resulted in significant improvements in energy savings and operational efficiency. This highlights the potential of GenAI to transform building energy management by providing accurate predictions and maintaining confidentiality of sensitive data.}
 

\subsubsection{3D City Modeling} 
\label{subsec:3d-city}
\rev{3D city models has played a pivotal role to support the 3D modelling and monitoring of buildings and infrastructure in urban areas} \citep{masoumi2023city}, \rev{and are also often to develop interactive and immersive visualizations of complex urban environments through virtual reality technologies.} These VR apps facilitate public engagement and collaborative urban planning \citep{dembski2019digital,xu2023toward}, simulation and evaluation of smart technologies \citep{xu2023semi,xu2023generative}, and 3D cadastral and \rev{architectural} design \citep{shojaei2013visualization, polys2018value}. 

Despite that the traditional geomatics and procedure-based methods \citep{lerma2010terrestrial,bejleri2021using} are capable of creating the Digital Surface Models (DSMs) of many large cities \citep{singh2013virtual}, these DSMs have limited resolution (e.g., texture and geometry) for representing building and infrastructure at the acceptable Level of Details (LoD). Additionally, many novel uses of 3D digital models (e.g., Hardware-In-the-Loop (HIL) simulations and scientific gamification) would require a gamified environment in which urban entities, such as buildings, roads, and infrastructure, are fully separated and have individual physical and gaming properties \citep{xu2023semi}. In this setting, creating a high-fidelity 3D representation of a city is a challenging effort that often requires massive, repetitive, time-consuming and manual efforts for surveying and 3D modeling of urban entities \citep{xu2023toward}, along with advanced computing resources for processing and integrating large-scale GIS, remote-sensing (e.g., photogrammetry and LIDAR scan), and crowd-sourcing data in a virtual environment \citep{xu2023semi}. It has been an emerging trend to leverage the 3D content-generation capability of the recent GenAI techniques to automate the tremendous manual efforts for creating 3D assets of urban entities, such as terrain, buildings, infrastructures, and vegetation \citep{xu2023semi}. 

\begin{description}
\item[3D Building Generation] The autonomous capability of recent GenAI models has allowed them to create photo-realistic 3D representations of buildings and street environments\citep{pang20223d,wu2022generative}. These applications have advanced the technical capability of many urban science fields, including architecture design, building construction and operations, and urban planning.
\citet{Du_3d_building_gan} introduces a novel method for 3D building fabrication using a Hybrid GAN. This method incorporates a unique loss function, including cycle consistency and perceptual loss, and utilizes a multi-properties GAN chain to create complex architectural designs. It also features a mixed GAN network for generating geometry and texture coordination. This approach significantly enhances the realism of 3D building models, achieving a 20\% performance improvement over traditional methods, demonstrating its potential in applications like digital city construction and smart urban planning. \citep{kelly2018frankengan} discusses an innovative framework called FrankenGAN. This system is designed to enhance coarse building mass models by adding detailed geometric and texture details, creating realistic and visually appealing urban landscapes. FrankenGAN employs a cascade of GANs, synchronized to maintain consistent styles across multiple scales in neighborhoods. It offers user control over the variability of output styles through exemplar images, allowing for a blend of automated and user-guided design. This technology is significant for its potential to revolutionize urban modeling and architectural design, providing a powerful tool for researchers and professionals in these fields to generate detailed, large-scale 3D city models efficiently and with high fidelity. \citet{bacharidis20203d} explores the use of GANs for reconstructing 3D building facades from images. It employs deep learning techniques for facade structural element detection, depth point-cloud generation, and protrusion estimation. The study demonstrates GANs' capability in segmenting building components from images and generating detailed 3D facades. These capabilities show GANs' potential in evaluating urban scenes and extracting urban design parameters, such as window ratio, skyline, and other applications, including cultural heritage preservation. The approach aims to address the challenges of architectural diversity and data scarcity in 3D modeling.

\item[Street Environment and Infrastructure]
A few other studies have developed GenAI strategy to digitize the entire street environment (with buildings and infrastructure) as 3D models, laying an foundation for the next-generation extended reality applications for supporting participatory and collaborator urban design and planning \citep{xu2023toward}. 
\citep{kim2020citycraft} introduces a novel method for generating three-dimensional virtual models of cities from a single street-view image. This approach differs from traditional reconstruction methods by utilizing machine learning, specifically GANs and convolutional neural networks (CNNs), to create terrain maps and identify city components and styles. The generated 3D virtual cities are visually plausible and resemble actual cities. This method represents a significant advancement in the field of urban modeling, offering a unique and efficient way to create detailed, large-scale city models, which is particularly useful for applications in gaming, film, and urban planning. The use of GANs and CNNs in this context is notable for its ability to support research and development in these areas by providing a more streamlined and user-friendly approach to city modeling.
\end{description}

\rev{In addition to GenAI models, many deep learning-based computer graphics and vision techniques have also been developed to facilitate the 3D reconstruction of urban environments using 2D images as inputs. We would like to discuss these applications as alternatives to popular GenAI models. Example of these techniques include Neural Radiance Fields (NeRF) \citep{mildenhall2021nerf} and more recent 3D Gaussian Splatting (3DGS) \citep{kerbl20233d}. NeRF represents a significant advancement in this field. By utilizing deep neural networks, NeRF facilitates the direct mapping of spatial coordinates to color and density\citep{chen2024survey}. Its success is rooted in its ability to create continuous, volumetric scene functions, resulting in outputs with unparalleled detail and realism. \citet{tancik2022block} presents a significant advancement in neural rendering for large-scale urban environments. It introduces Block-NeRF, a variant of Neural Radiance Fields (NeRF), specifically designed to handle city-scale scenes. This method effectively decomposes a large scene into individually trained NeRFs, allowing for scalable rendering and enabling updates to specific parts of the environment without retraining the entire network. This technique represents a major step forward in the field of photorealistic 3D reconstruction and neural rendering. }

\rev{Despite the advancements, NeRFs still face the challenge of slow rendering speeds and visual quality due to the computationally intensive nature of volume rendering\citep{barron2022mip, chen2022tensorf}. To address these limitations, 3DGS was developed and has been applied to efficiently reconstruct large-scale 3D urban environment \citep{chen2024survey}. This technique explicitly models scenes using 3D Gaussians with learnable attributes and performs rasterization of these Gaussians to produce renderings \citep{kerbl20233d}. \cite{cheng2024gaussianpro} presents a new method for rendering extensive 3D scenes efficiently. By introducing a hierarchical structure of 3D Gaussians, the technique enables efficient level-of-detail (LOD) rendering, optimizing visual quality and speed. The method involves dividing the scene into smaller chunks for parallel processing, reducing resource demands. Adaptations for sparse data and optimized training ensure high visual fidelity and smooth transitions between detail levels. This approach supports real-time rendering of large scenes using affordable equipment, making it practical for applications like virtual reality and urban planning. The authors demonstrated the method's capability to handle tens of thousands of images over several kilometers.
\citep{liu2024citygaussian} introduces CityGaussian (CityGS), a method to efficiently render large-scale 3D scenes in real-time using 3D Gaussian splatting. The authors address the challenges of training and rendering large-scale scenes by employing a divide-and-conquer strategy and a Level-of-Detail (LoD) approach. The method begins with generating a global Gaussian prior, followed by partitioning the scene into manageable blocks, which are then trained and refined independently. To enhance rendering efficiency, CityGS uses block-wise LoD selection to feed only necessary Gaussians into the rasterizer, significantly reducing computational load while maintaining visual quality. Experimental results demonstrate that CityGS achieves state-of-the-art rendering quality and real-time performance across various large-scale scenes, outperforming existing methods in both visual fidelity and speed. 
\citep{lin2024vastgaussian} presents VastGaussian, a method for reconstructing large 3D scenes with high fidelity and real-time rendering using 3D Gaussian splatting. The authors address the limitations of existing methods, such as visual quality and rendering speed, by proposing a progressive partitioning strategy that divides scenes into multiple cells. Each cell is optimized independently and then merged to form a complete scene, which allows for efficient memory usage and faster optimization. They also introduce a decoupled appearance modeling technique to handle appearance variations, ensuring seamless integration and reducing floaters in the rendered output. VastGaussian outperforms existing NeRF-based methods in both quality and efficiency, achieving state-of-the-art results on several large scene datasets.}

 
\subsection{Urban Planning} 
\label{subsec:urban-design}
\rev{Out of the 201 articles chosen for their relevance to urban planning, we conducted a further screening based on titles, abstracts, and keywords. This process enabled us to pinpoint the primary GenAI use cases suitable for integration into urban digital twins, especially within the realms of urban population synthesis, urban morphology analysis, land use and land cover management, urban design, and urban disaster management. The detailed use cases are discussed in the following subsections.}

\rev{
Urban digital twins are advanced virtual models of cities that integrate real-time data and simulations to enhance urban planning and management, encompassing various urban sybsystems and their interaction from a holistic and sociotechnical perspective \citep{schrotter2020digital}. Previously, urban planners and researchers have utilize urban digital twins as powerful data-driven tools to visualize and analyze complex urban dynamics, conducting predictive analysis, decision optimization, and resource allocation \citep{al2021digital}. In additional, the digital replica of cities have been frequently used as the playground to facilitate the testing of hypothetical planning scenarios and urban designs, such as land use and land cover planning and street layouts. The integration of these digital tools in urban planning fosters more resilient, sustainable, efficient, human-centered, and livable urban environments \citep{xia2022study, schrotter2020digital}. Many previous studies in the urban planning sector have highlighted that the capabilities of GenAI models in creating realistic, high-dimensional data and hypothetical planning scenarios based on historical data can be leveraged to mimic complex urban dynamics and processes, thereby facilitating urban analysis. By creating synthetic data and scenarios of various urban growth patterns, demographic distributions, and city layouts. 
}. 

\rev{Motivated by the interest in integrating GenAI with digital twins, we have structured this section to highlight the key functionalities of urban digital twins in urban planning. This is organized through the following subsections. Additionally, we have selected and reviewed representative GenAI applications in urban planning that align with the data-driven focus of urban digital twins, specifically tailored to support major urban planning tasks.
}

\subsubsection{\rev{Urban Population Synthesis} }
\rev{Urban population synthesis is crucial to urban planning as it provides accurate demographic data, enabling planners to design equitable services and infrastructure. It supports informed decision-making, sustainable development, and efficient resource allocation, ensuring that the diverse needs of the urban population are met \citep{bibri2021data}.}
\rev{
\citet{garrido2020prediction} explores the application of GenAI models, specifically using Wasserstein Generative Adversarial Networks (WGAN) and VAE, to address challenges in urban population synthesis. The study focuses on predicting rare combinations of features in high-dimensional datasets, which traditional methods struggle with due to the curse of dimensionality and sparsity of survey data. By employing deep generative models, the research demonstrates significant improvements in accurately generating synthetic populations that maintain realistic joint distributions of multiple variables. This approach effectively recovers missing data (sampling zeros) while minimizing impossible combinations (structural zeros), enhancing the accuracy and scalability of population synthesis for urban planning and agent-based modeling applications.
\citet{johnsen2022population} discusses the use of deep generative models, specifically Conditional Variational Auto-Encoders (CVAE) and Conditional Generative Adversarial Networks (CGAN), to model the population distribution for new real estate developments. The study utilizes real data from Ecopark Township in Hanoi, Vietnam, to create synthetic populations that mirror the demographic characteristics of actual residents. These models help urban planners predict the impact of new developments by generating realistic synthetic populations based on property-specific attributes. The research highlights the superiority of the CVAE model in generating accurate and scalable population data, demonstrating the potential of GenAI to enhance urban planning and decision-making processes. \citet{aemmer2022generative} explores the use of GenAI, specifically VAE and Conditional Variational Autoencoders (CVAE), to synthesize urban populations for planning purposes. This method addresses the limitations of traditional population synthesis techniques, which struggle with scalability and accuracy when handling numerous variables and geographic regions. The proposed VAE/CVAE model can generate synthetic populations that accurately represent both household and individual characteristics, even with sparse training data. The study demonstrates that the VAE/CVAE model outperforms traditional methods in terms of accuracy and computational efficiency, particularly when dealing with complex, multi-dimensional datasets. This innovative approach enhances urban planning by providing detailed and scalable population data, crucial for microsimulation models and transportation network development.
}

\subsubsection{\rev{Urban Morphology} }
\rev{Urban morphology analysis is crucial to urban planning as it provides insights into the physical structure and spatial patterns of cities, enabling planners to design more efficient, sustainable, and livable urban environments. GenAI models play a transformative role in urban morphology analysis by automating data processing, enhancing simulation capabilities, and urban system optimizing\citep{yigitcanlar2024artificial}. \citet{zhang2022metrogan} introduces Metropolitan GAN (MetroGAN), a novel Generative Adversarial Network framework, for simulating urban morphology with geographical knowledge. MetroGAN addresses limitations of previous GAN models such as data sparsity and training instability by incorporating a progressive growing structure and a geographical loss for water area constraints. It significantly outperforms existing urban simulation methods, demonstrating over 20\% improvement in all metrics. Interestingly, MetroGAN can generate city shapes using solely physical geography features, showcasing its ability to handle various urban attributes and stabilize GAN-based urban simulation. \citet{allen2022generative} delves into the application of deep learning, particularly GANs, for analyzing urban morphology. This research highlights the capability of GANs to process and interpret complex urban data, providing insightful analyses of city structures and patterns. The innovative use of GANs in this context offers significant contributions to urban planning and development, enhancing our understanding of urban spaces and their evolution. }

\subsubsection{\rev{Land-use and Land-cover Management} }
\rev{Land-use and land-cover management is crucial to urban planning as it ensures sustainable development, efficient infrastructure use, and disaster risk reduction. It promotes economic growth, social equity, and legal compliance by guiding urban growth, preserving natural resources, and integrating urban planning \citep{kalfas2023urbanization}. }

\citet{sun2021gan} explores land-use and land-cover change (LUCC) prediction using deep learning. It introduces a model employing GANs, specifically the pix2pix GAN, combined with an attention mechanism. This approach significantly enhances local spatial detail and diversity in urban planning and LUCC prediction. The model's unique capability lies in fusing multiscale local spatial information with city planning data to predict future land-use changes, showcasing a novel application of GANs in urban planning and environmental monitoring. 
\rev{\citet{he2022use} explores the application of GenAI in urban morphology analysis. It introduces the UNET-GAN-CD model, which combines the strengths of UNET and GANs to enhance change detection in urban environments. The model processes bi-temporal satellite images to identify changes in urban landscapes, such as road modifications, land use alterations, and building constructions. By automating data processing and utilizing advanced simulation capabilities, the UNET-GAN-CD model significantly improves the accuracy and efficiency of urban change detection, facilitating better-informed urban planning decisions. The study highlights the transformative role of GenAI in optimizing urban system analysis, making it a valuable tool for sustainable and efficient urban development.}

\subsubsection{Urban Design} 
Generative design is defined as using computer-aided generative methods to produce design options, practices, and solutions, improving certain stages of the typical design process \citep{stojanovski2020city, buhamdan2021generative}.
In recent years, GenAI models play a crucial role as the intelligent agent in computer-aided urban design. They are employed in many studies of urban design alternatives and strategies that are driven by urban big data and previous solutions \citep{al2021generative}. 

Recent progress in generative urban design is evident in the urban science and smart city sector, where these models are increasingly used for tasks such as urban layout and land use optimization, environmental impact assessments, and urban resource management. 
At a high level, a study by \citet{quan2019artificial} highlighted the effectiveness of AI in managing urban complexity and sustainability, as well as supporting generative and participatory urban design. Furthermore, the integration of AI in smart cities is demonstrated by the work of \citet{batty2018artificial}, who emphasized AI's role in analyzing large-scale urban data to inform better city planning and management. The trend towards AI-aided urban planning signifies a paradigm shift, promising smarter, more efficient, and sustainable urban environments.

As examples of detailed generative urban design applications, \citet{huang2022accelerated} explores the use of GANs, in urban design for environmental performance optimization. It demonstrates the use of GANs in rapidly generating urban design alternatives that meet specific environmental criteria, highlighting deep learning's ability to process complex spatial data and simulate various urban development scenarios. This approach represents a significant innovation in integrating environmental considerations into the urban design process using advanced AI techniques.
\citet{fedorova2021generative} examines the use of deep learning, particularly GANs, in creating urban layout designs. It focuses on GANs' ability to generate diverse and realistic urban layouts, which can significantly aid in urban planning and development. This research underscores the potential of deep learning in addressing complex spatial design challenges, providing innovative tools for urban designers and planners.
\citet{noyman2020deep} explores the application of GenAI models, such as GANs and VAEs, in urban design, focusing on generating urban block layouts. This innovative approach does not require predefined parameters for design solutions, as the neural network learns the visual features of existing city structures. The research demonstrates GANs' ability to adapt designs to various urban contexts and morphologies. Significantly, it highlights the potential of GenAI models in urban planning, offering a novel method for designing urban blocks that structurally align with the existing urban fabric without the need for explicit parameter definitions. It should be mentioned that social, economic, and environmental considerations can be used as constraints to improve the urban blocks from the livability and sustainability perspectives.
\citet{quan2022urban} develops Urban-GAN, an AI-aided generative urban design computation system, which empowers public participation in urban design through GANs and Deep Convolutional Neural Networks (DCNNs). This innovative system enables individuals with little design expertise to generate professional-level urban designs. It employs sub-symbolic representation and case-based reasoning, facilitating users to select urban form cases and generate similar design schemes. The system's application in hypothetical design experiments across cities like Manhattan, Portland, and Shanghai illustrates its effectiveness in generating diverse urban forms, highlighting its potential in democratizing urban design and planning.
\citet{kimparticipatory} focuses on developing an AI tool called PlacemakingAI, which uses GANs for urban design. The tool facilitates public participation in urban planning, allowing users to visualize sustainable urban spaces in real-time. It leverages both supervised and unsupervised GANs to generate synthetic images from datasets of walkable streets. This research is significant for its innovative use of GANs in urban design, enabling a participatory approach and fostering engagement from citizens, designers, and stakeholders in the urban design process.

\rev{We discovered a very recent study that utilizes diffusion models in the generative urban design sector. \citet{cui2024learning} highlights the use of GenAI, specifically a latent diffusion model, to create high-quality urban design images. The study addresses the challenges of precise layout control in image generation by integrating a conditional control network and employing a low-rank adaptation strategy to manage the training complexity. By leveraging these generative techniques, the model produces realistic urban design images based on text prompts and line art inputs. The proposed framework demonstrates significant improvements in generating detailed and controlled urban design layouts, enhancing the practical applications of AI in urban planning and design.}

\subsubsection{Disaster Management}
Generative models, adept at synthesizing realistic images and scenarios, are being utilized for detecting particular disaster and simulating potential disaster impacts, enhancing preparedness strategies. In earthquake and flood management, for instance, GANs generate detailed images of damaged infrastructure, aiding in rapid damage assessment and efficient resource allocation \citep{tilon2020post}. This technology significantly improves response times and decision-making accuracy, proving vital in mitigating the effects of urban disasters and bolstering resilience in city planning and emergency responses. 

As detailed examples, \citet{hofmann2021floodgan} presents a novel deep learning approach for real-time pluvial flood prediction. The study introduces floodGAN, a deep convolutional generative adversarial network, which significantly accelerates flood prediction while maintaining high accuracy. This approach effectively utilizes the capabilities of deep learning, particularly GANs, to understand and predict complex flood dynamics based on spatially distributed rainfall events. The innovation lies in its ability to provide fast, accurate flood predictions essential for real-time applications such as early warning systems, marking a substantial advancement in flood modeling and risk management. \citet{salih2022comparison} explores the use of GANs and Mask Region-Based Convolutional Neural Networks (Mask-R-CNN) for floodwater detection and segmentation in images. The study highlights the capability of GANs to generate segmented images of flooded areas, emphasizing their potential in applications where quick and accurate floodwater identification is crucial. The research demonstrates the effectiveness of GANs in image-based flood detection, offering a novel approach to enhancing flood response and management strategies. \citet{yadav2024unsupervised} introduces an innovative unsupervised approach for flood mapping using Synthetic Aperture Radar (SAR) time series data. It leverages a VAE trained with reconstruction and contrastive learning techniques. This model is adept at detecting changes by examining the latent feature distributions between pre-flood and post-flood data. Its performance, evaluated against nine different flood events, demonstrates superior accuracy over existing unsupervised models. This approach marks a significant advancement in the application of deep learning to environmental monitoring, specifically in enhancing flood detection capabilities using satellite data. \citet{lago2023generalizing} introduces cGAN-Flood, a conditional generative adversarial network developed to enhance flood prediction in urban catchments. Unlike traditional ANN models, cGAN-Flood can generalize flood predictions to areas outside its training dataset, addressing varying boundary conditions. It achieves this by using two cGAN models for identifying wet cells and estimating water depths, trained with HEC-RAS outputs. The method was tested on different urban catchments and compared with HEC-RAS and the rapid flood model WCA2D. It successfully predicted water depths, despite some underestimation in channels during low-intensity events. Significantly faster than existing models, cGAN-Flood offers a promising, efficient alternative for large-scale flood forecasting.
\citet{delacruz2020using} presents an innovative approach using GANs for rapid classification of earthquake-induced damage to railways and roads. It highlights the use of GANs to generate synthetic images of damage, addressing the challenge of limited real-world data for training deep learning models. This method demonstrates the potential of GANs in enhancing the efficiency and accuracy of structural damage assessment, which is crucial for emergency response and infrastructure recovery efforts. \citet{zhou2023automatic} utilizes GANs for urban morphology prediction related to local climate zones (LCZs). It employs GANs to integrate LCZ and urban morphology data from six cities, demonstrating superior accuracy and convergence capabilities, particularly with the Pix2pix model. This approach allows for rapid and responsive 3D urban morphology prediction, offering significant insights for enhancing local climates in urban areas through 3D generative design techniques.

%% file: Content/40-future-discussion.tex
\section{Future of \rev{GenAI} in Urban Digital Twins} 
\label{sec:Future Discussion}
\rev{To address research question Q4, proposed in Section \ref{sec:research-questions}, regarding how GenAI technologies can benefit urban digital twins, this section explores the integration of GenAI models with existing urban digital twin applications (e.g., web-based platforms, cyber-physical systems, cyberinfrastructure, and visual dashboards). The aim is to augment their current capabilities for more efficient participatory smart city management through human-AI collaboration. Additionally, we delve into the potential technical and methodological challenges associated with the development of generative urban digital twins, underscoring the complexities and innovations required for this visionary approach.} 

\begin{figure*}[htbp]
 \centering
\includegraphics[width=\textwidth]{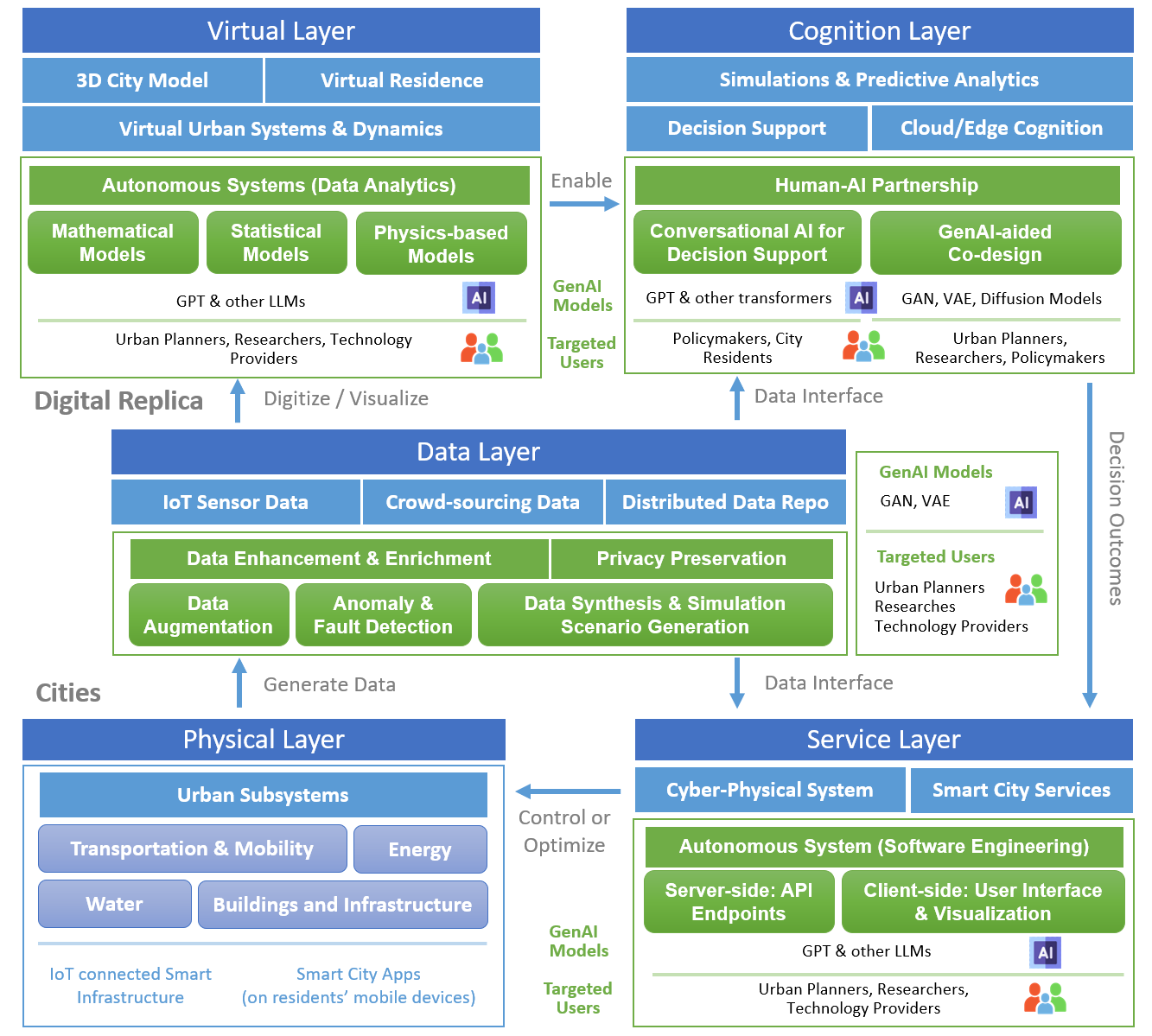}
 \caption{\rev{Visions and opportunities of GenAI models} for enhancing cognitive urban digital twins at the conceptual level.}
 \label{fig:dt-opportunities}
\end{figure*}

\subsection{\rev{Cognitive Digital Twins: Visions and Opportunities}\label{sec:cdt-vision}} 
As an integration between AI and digital twins, \rev{the emerging} Cognitive Digital Twins \rev{(CDTs)} represent a significant advancement in the digital twin concept, \rev{enabling self-learning, memory, and reasoning capabilities beyond their traditional counterparts \citep{niloofar2023general}. } These \rev{AI} models assimilate cognitive functions, enabling CDTs to dynamically process and learn from data. They harness deep learning to analyze complex datasets, predict future scenarios, and make decisions autonomously, leading to improved efficiency and adaptability \citep{ali2021cognitive}. In urban science and smart city applications, \rev{the CDT concept illuminates more intelligent, efficient, and automated solutions to complex urban management problems by leveraging the advanced capabilities of various neural networks for classification, fault detection, object detection, and optimization.}

\rev{Recent revolutionary advancements in GenAI can be integrated into CDT to further enhance its capability for more efficient and participatory urban management in the aspects of (a) facilitating data enhancement and enrichment, (b) building autonomous systems, and (c) promoting human-AI partnership. We present our vision of how GenAI models could be integrated into CDT for smart cities through Figure \ref{fig:dt-opportunities}, in which we adopted the core layers of a CDT defined by previous studies \citep{shi2022cognitive,abburu2020cognitwin} and elaborated on who individual layers could be benefited from GenAI through the following subsections. The essential layers of a CDT are represented by dark blue boxes, while our proposed visions and opportunities are depicted in green boxes in the figure.} 

\subsubsection{\rev{Data Enhancement and Enrichment} }
\rev{Data quality and utility is critical to data-driven applications, especially urban digital twins that leverage urban data to enable the situational awareness, predictive analytics, and informed decision supports of various urban subsystems. 
\begin{description}
 \item[Data Enhancement] refers to the augmentation of existing data sets by adding more detailed information, often from external sources, to make the data more comprehensive. In this context, GenAI models, such as GANs and VAEs, have been widely utilized in contemporary studies to generate realistic data \citep{wu2022generative}, thereby enhancing the reliability and accuracy of a CDT's situational awareness capabilities. Additionally, VAEs are particularly effective in facilitating fault and anomaly detection, which further supports data enhancement efforts \citep{guo2020unsupervised}. VAEs operate by encoding input data into a latent space and then decoding it back to the original space, a process that enables them to capture complex data distributions and identify deviations from normal patterns. This ability to detect anomalies and generate synthetic data contributes to creating more robust and comprehensive datasets, ultimately improving the quality and utility of the data.  
 \item[Data Enrichment] involves the integration of additional data to provide more context, making the data more meaningful and useful for analysis. In this regard, GenAI models, such as GANs and VAEs, have been frequently utilized in urban studies to generate synthetic data and simulate scenarios \citep{figueira2022survey}. 
\end{description}
The data synthesis capabilities of these models enhance an urban digital twin's predictive analytics, allowing for accurate projections and forecasts of urban dynamics in futuristic or hypothetical management scenarios. For instance, the design and development of critical infrastructure, such as airports, and their complex interactions with urban transportation systems can be modeled and simulated through various scenarios. By integrating proposed future developments (e.g., new buildings and transport networks) with current and real-time data (e.g., traffic patterns, existing buildings, and land uses), these models enable the creation of "what-if" scenarios. These scenarios provide digital twin users with alternatives for more adaptive and informed decision-making, ultimately supporting more resilient and efficient urban planning and management. By leveraging GenAI, urban digital twins can achieve more comprehensive, context-rich data, ultimately leading to more reliable data-driven insights and inferences, as well as more accurate predictive analytics.}

\rev{This enhancement creates opportunities for making better-informed urban decisions and strategies, providing effective solutions to address digital twin challenges C1 and C2 discussed in Section \ref{sec:udt-challenges}.}

\subsubsection{\rev{Privacy-Preserving Synthetic Data Generation}}
\rev{Privacy-Preserving Synthetic Data Generation refers to the use of advanced generative AI techniques, such as GANs and VAEs, to create artificial data that closely mimics real-world data while safeguarding the privacy of the individuals whose information is being represented \citep{liu2022privacy, ghatak2022survey}. This approach is particularly crucial in urban research, where the collection and utilization of extensive datasets, such as mobility trajectories and building occupancy data, pose significant privacy and sovereignty concerns.} \rev{In urban research, mobility trajectory data, which tracks individuals' movements across a city, can reveal sensitive information about their daily routines, frequently visited locations, and social interactions. Similarly, building occupancy data can disclose patterns related to the presence and activities of individuals within specific buildings, raising potential privacy risks. By generating synthetic data that replicates the statistical properties and patterns of these datasets, researchers can conduct analyses and derive insights without exposing personal information \citep{abay2019privacy}.}

\rev{Generative AI models are instrumental in this process due to the advantages provided by their architectures. GANs consist of two neural networks—a generator and a discriminator—that work in tandem to produce data indistinguishable from real data. The generator creates synthetic data, while the discriminator evaluates its authenticity. Through iterative training, GANs improve the quality of the synthetic data, ensuring it accurately reflects real-world patterns.
VAEs, on the other hand, encode real data into a latent space and then decode it back into synthetic data, preserving the essential characteristics while obscuring individual-specific details. This approach ensures that the generated data maintains the integrity and variability of the original data while enhancing privacy and reducing the risk of re-identification.
By employing these generative AI techniques, urban researchers can create privacy-preserving synthetic datasets that enable robust analysis and innovation. This method not only addresses ethical and legal concerns related to data privacy but also enhances the availability of high-quality data for urban planning, policy-making, and smart city development.} 

\rev{In regards, privacy-preserving synthetic data generation can effectively address the challenges posed by data ownership, data residency, data colonialism by creating artificial datasets that mimic real data without containing any actual sensitive and protected information \citep{boedihardjo2022privacy}. This approach allows organizations to comply with local data residency laws and facilitates ethical cross-border data sharing, mitigating data colonialism by enabling ethical data sharing \citep{majeed2023attribute, vie2022privacy}. Additionally, it empowers local stakeholders by enabling them to generate and control representative datasets, thus reducing dependency on external entities and promoting fair data practices ,and providing effective solutions to address digital twin challenge C5 discussed in Section \ref{sec:udt-challenges}. }

\subsubsection{\rev{Autonomous Systems} }
\rev{Recent developments in GPT and RAG have highlighted the potential for creating autonomous systems or agents to automate both data analytics and software development \citep{suri2023software}. Building an urban digital twin requires significant interdisciplinary efforts in data management, analysis, and software engineering. In this regard, integrating \rev{GenAI} models, particularly GPTs, can help automate labor-intensive, time-consuming, and repetitive tasks, thereby reducing the cost of developing and maintaining an urban digital twin. GenAI-powered autonomous systems can be deployed in the virtual and service layers of a CDT to generate computer code in various programming languages, facilitating the creation of core data analytics and software components.
}
\rev{
\begin{description}  
 \item[Autonomous Data Analytics] leverage GPTs' reasoning and generative capabilities to develop automated data workflows and pipelines that can perform advanced data analysis (e.g., statistical, mathematical, and spatial analysis) by calling appropriate libraries based on user-defined prompts \citep{vemprala2023chatgpt}. These autonomous systems often employ prompt engineering, predefined prompt bases and templates, and fine-tuning techniques to instruct large language models on specialized data analysis tasks. An example of an autonomous spatial analysis tool powered by the ChatGPT API is presented by \cite{li2023autonomous}. With the recent advancements in RAG, GPTs are now empowered with the capability to query external information sources through vector databases, broadening the application and utility of these generative models.   
 \item[Autonomous Software Development]  focuses on utilizing GPTs' code generation capabilities to develop critical software components \citep{suri2023software}. Numerous studies have explored GPTs' ability to create web applications, knowledge graphs, and databases using prompt templates and fine-tuning techniques \citep{liukko2024chatgpt,jamdade2024pilot,vamsi2024human}. Since urban digital twins are essentially web-based or cloud-based applications, their core software components could potentially be generated using GPTs and other pre-trained language models. Consequently, we envision an autonomous system for code generation as a component of the service layer in CDTs, designed to automate the time-consuming software development and maintenance tasks. The approach can generate codes for both the server-side and client-side applications. 
\end{description}
}
\rev{
Integrating GenAI-powered autonomous systems (i.e., agents) into a CDT can significantly lower the technical barriers and programming skills required for developing an urban digital twin. In terms of new application opportunities, this paradigm can reduce the costs associated with software development and maintenance while increasing the efficiency of domain experts, such as urban planners and researchers in academia with moderate coding experience, allowing them to develop robust urban digital twin applications. These autonomous systems facilitate the creation of specialized research software, including digital twins, decision support systems, and content management systems. Consequently, researchers can focus more on their domain expertise and less on technical implementation and software engineering practices.}\rev{This paradigm enable cost-effective solutions to address digital twin challenges C3 and C4 discussed in Section \ref{sec:udt-challenges}.}

\subsubsection{\rev{Human-AI Partnership} }
\rev{In the CDT's cognition layer, where human involvements are often required to make critical decisions, GenAI models and their generative abilities can be harnessed to promote human-AI partnerships for a more participatory, adaptive, collaborative urban management approaches, which match the demanded and skills of various participants in urban managements. These participants include domain experts (e.g., urban planners and researcher), policymakers, and city residents. The proposed approaches are centered are around two aspects, namely conversational AI for decision support and GenAI-aided co-design.
} 
\rev{
\begin{description}  
 \item[Conversational AI] refers to emerging AI agents powered by advanced large language models, such as GPT, which enable intuitive and interactive communication between human users and AI systems \citep{mctear2022conversational}. Unlike traditional expert systems that often require users to possess domain knowledge, conversational AI offers a more natural, intuitive, and user-friendly interface. This allows individuals to participate in decision-making processes informed by data-driven insights, domain-specific simulations, and AI-powered predictive analytics \cite{sakirin2023user}. Conversational AI paradigms have proven effective in facilitating public engagement and participation in urban planning and management \citep{sahab2024conversational}. By providing a GPT-powered medium with excellent language understanding capabilities, these systems eliminate the need for users to have domain expertise. This approach enables interactive, user-centered dialogues with city residents and policymakers, capturing their needs and providing education and information relevant to support informed decisions. By integrating conversational AI into urban digital twins, GenAI promotes a more interactive, efficient, and democratic decision-making process, ultimately leading to more resilient and adaptive urban environments.  
 \item[GenAI-aided Co-Design] can revolutionize urban planning and management by enabling more dynamic, collaborative, and innovative approaches to city development and management. GenAI models assist in co-design processes by interpreting and responding to user inputs, allowing stakeholders to collaboratively explore various planning scenarios and design alternatives. Leveraging the generative capabilities of models such as GANs, VAEs and generative diffusion models, trained on successful design blueprints and practices, domain experts (e.g., urban planners and researchers) can automate the creation of urban designs (e.g., floor plans, street layouts, landscapes, and parks) across large geographic regions in a short period of time \citep{guridi2024image, quan2022urban}. This paradigm not only lowers costs but also enables urban planners to formulate design problems and explore more options through AI-Generated Content (AIGC) \citep{jiang2023generative}. Furthermore, GenAI-powered tools facilitate co-design processes by allowing stakeholders, including community members, policymakers, and architects, to interact with AI systems through intuitive interfaces. This interaction can include generating and visualizing different design alternatives, optimizing layouts for sustainability and efficiency, and integrating inquiries and feedback in real-time \citep{knight2023generative}.
 \end{description}
}
\begin{figure*}[htb]
 \centering
\includegraphics[width=\textwidth]{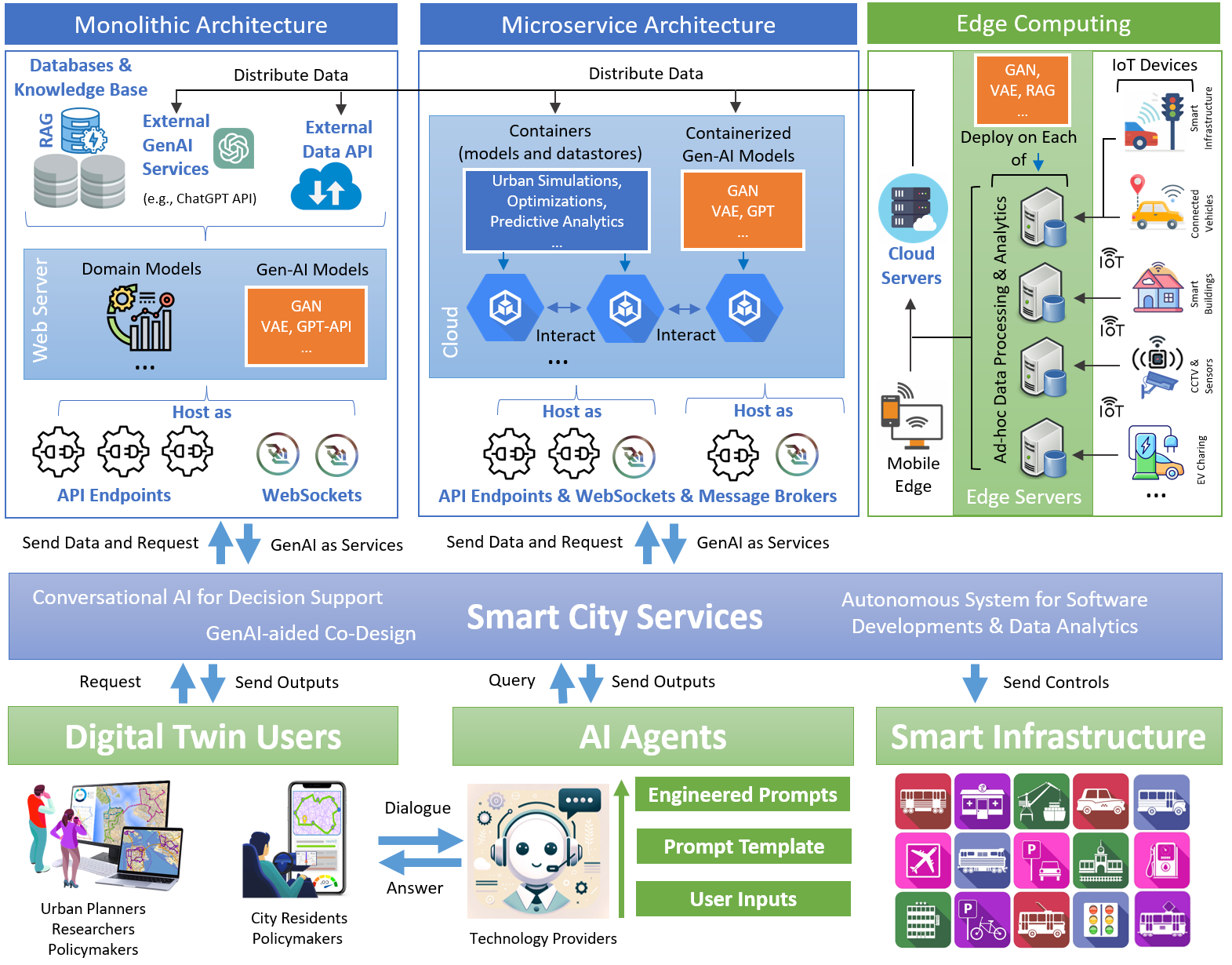}
 \caption{Technical strategies for AI-Digital Twin integration: Generative \rev{(GenAI)} models deployment in different digital twin architecture. }
 \label{fig:dt-ai-integration}
\end{figure*}
\rev{By integrating GenAI models into urban digital twins, we can create opportunities for generating more inclusive, adaptable, and responsive solutions to the evolving needs and values of various urban management participants. These generative approaches not only enhances the quality and sustainability of urban environments but also empowers communities to play an active role in shaping their future by building transparent and trustworthy human-AI partnership. This approach could provide efficient solution to address challenge C6 discussed in Section \ref{sec:udt-challenges}.}

\subsection{GenAI-Digital Twin Integration: \rev{Technical Strategies} }
\rev{This subsection explores potential technical implementation strategies for integrating popular GenAI models into existing urban digital twin applications.}
Through the evolution history of the urban digital twins, a few software architectures emerge as particularly adept at fostering sophisticated, scalable, and efficient digital representations of urban landscapes. Among these, monolithic architecture, microservices, and edge-computing architecture are pivotal \citep{ferko2022architecting}. Monolithic architecture lays the groundwork by facilitating interoperable services that communicate across a network and is crucial for developing basic cyberinfrastructure for amalgamating diverse simulation systems and data sources using the traditional server-client paradigm \citep{gos2020comparison}. Evolving from monolithic architecture, microservices architecture advances this model by organizing applications into collections of loosely coupled services that are often adopted in large-scale cloud-based applications. This structure promotes agile development and deployment of complex cyberinfrastructure systems, enabling seamless updates and scalability for specific elements (e.g., simulation models for specific urban subsystems) of an urban digital twin without impacting the overall framework \citep{ponce2019migrating,xu2023smart}. 
Meanwhile, Edge computing introduces a move towards decentralized and distributed data processing, allocating computational tasks nearer to the data origins at the network's edge nodes \citep{yu2017survey}. Examples of these edge nodes are often edge servers that are directly connected to sensors and mobile IoT devices for sensing various types of urban data \citep{xu2022mobile,marjanovic2018edge}. This shift is vital for urban digital twin development, as it allows for real-time data analysis and immediate response critical for the effective management of dynamic urban activities. Collectively, these architectural strategies create a formidable foundation for constructing urban digital twins, delivering the adaptability, scalability, and immediacy required to simulate and manage complex urban systems accurately.

To effectively integrate \rev{genAI-powered applications} within urban digital twins—often manifested through web-based platforms, cyberinfrastructures, and cloud computing services for enhanced smart city functionalities—we propose several technical strategies (as demonstrated in Figure \ref{fig:dt-ai-integration}). These strategies are designed to bolster scalability, performance, and precision in the AI-digital twin synergy. Tailored specifically to accommodate urban digital twins, these approaches are aligned with distinct architectural patterns. This ensures that each strategy not only addresses the unique challenges of integrating \rev{GenAI models} into these digital twins' complex systems but also leverages the inherent strengths of the chosen architectural framework. By doing so, our approach facilitates the creation of dynamic \rev{modular, and maintainable} urban digital twins that can evolve with the changing needs of smart city ecosystems. \rev{From the technical perspective, we propose the following strategies:}
\begin{description}
    \item[\rev{GenAI} as a Service:] To integrate AI models into digital twins that adopt monolithic architecture and microservices architecture, a reliable practice would be hosting these models as web services following the paradigm of the service-oriented architecture \citep{schnicke2020enabling}. Currently, many cloud computing platforms provide the foundational infrastructure needed for deploying ML and DL models as web services at scale, offering flexible resources and cyber-deliver capability that can dynamically adjust to the computational demands of \rev{GenAI}. Utilizing the recent containerization technologies, such as Docker, and orchestration tools like Kubernetes can further enhance scalability and manageability by allowing AI components to be deployed as microservices. This modular approach facilitates the seamless integration of \rev{GenAI} models into existing urban digital twin frameworks, ensuring that complex simulations and data analyses run efficiently across distributed computing environments.
    
    \item[Ad-hoc Data Processing and Analytics at Edge Nodes:] For urban digital twin applications that involve real-time data streaming and simulation capabilities, deploying \rev{GenAI} models on the edge nodes can reduce latency and bandwidth use by processing data closer to the data source (e.g., sensors and IoT devices), which is crucial for real-time applications like traffic flow optimization, building and grid operations, or emergency response simulations within urban digital twins. \rev{GenAI} models can be readily deployed on edge servers or data loggers that are directly connected to urban sensors for processing and distributing raw data. This practice is complemented by the use of data streaming and real-time analytics platforms, such as Apache Kafka and Apache Spark, which can handle the high-velocity data feeds sensed from urban environments and provide the necessary processing power to augment and enrich urban data for supporting enhanced situational awareness and simulation-based predictive analytics in real-time.
     \rev{\item[GPT Prompt Base and Templates for Smart Cities:] The emerging concept of GPT Prompt Template Base involves developing structured templates and foundational setups that guide Generative Pre-trained Transformers (GPT) in generating accurate and contextually relevant outputs. These templates include predefined patterns and placeholders that can be filled with specific information, ensuring consistency and efficiency in prompt creation. By facilitating the autonomous generation of data and code, GPT Prompt Template Bases can significantly streamline the development of urban digital twins. They enable domain experts, such as urban planners and researchers, to easily generate complex data analyses, predictive models, and software components for an urban digital twin without extensive programming expertise. This is particularly important for urban sciences and management, where the integration of diverse datasets and the need for rapid, adaptive decision-making are critical. Creating robust GPT Prompt Template Bases for these fields ensures that AI-generated content is reliable, relevant, and aligned with the specific needs of urban development and management, ultimately enhancing the effectiveness and scalability of urban digital twin applications.}     
\end{description}

\rev{
We envision that through the proposed AI-digital twin integration, next-generation urban digital twins could provide more intelligent, intuitive, and straightforward interfaces to deliver personalized smart city services to various IoT-connected devices. These devices include mobile devices (e.g., smartphones, tablets, laptops, and desktops), which host CDT software (e.g., web apps or native apps) as AI agents to directly deliver smart city services to users. Additionally, smart infrastructure can sense real-time urban data and receive cyber-physical controls from the urban digital twins to optimize real-world urban systems. We present several potential use cases that would benefit from the proposed GenAI-digital twin integration:
(a) A smart city service delivered through an AI agent that allows city residents to request the generation and visualization of new park designs or building renovation plans, visualized through a mobile augmented reality environment rendered via the user's smartphone camera view. (b) A smart city service deployed on edge devices to improve the density and quality of traffic data collected through an IoT-connected Radar Detection System (RDS) sensor. This service delivers the enhanced and enriched RDS data to the urban digital twin's computing environment, providing urban planners with more accurate situational awareness of traffic flow and lane occupancy on urban roads.   
}

\section{Challenges and Limitations} 
Although numerous \rev{GenAI applications} have been developed to address challenges across various sectors of smart cities, significant obstacles persist. These challenges often arise from the limitations of GenAI models from both data science and computational perspectives \citep{duan2019artificial}, as well as from socio-ethical issues related to the AI-digital twin integration \citep{filipenko2024application}. \rev{In this regard, we categorize these challenges based on their nature in the following subsections and provide a detailed discussion of each challenge, along with potential solutions.}

\subsection{\rev{Data Science Challenges}}
\rev{The data science challenges of GenAI applications refer to the significant demands on computational resources required for training, deploying, and maintaining these models, including high processing power, extensive memory, and energy consumption, which can impact both cost and environmental sustainability \citep{wang2023overview}.}
\begin{description}
\item[Inherent Model Instability:] Generative AI models, such as GANs, often face difficulties in the data training processes to make these models reach a stable state where the generator and discriminator are well-balanced. This instability can manifest in several ways, including mode collapse, non-convergence, vanishing gradients, hyper-parameter sensitivity, and divergence \citep{sajeeda2022exploring,saxena2021generative}.
\item[Computing, Data Storage, and Latency:]
In the context of smart cities and urban digital twin applications, \rev{GenAI} introduces substantial computational challenges, necessitating high processing power to facilitate complex simulations of urban systems and real-time data analytics \citep{karapantelakis2024generative}. Moreover, these applications demand rapid and efficient data storage solutions to handle the extensive data volumes generated and processed, encompassing everything from sensor data to detailed environmental models. The seamless integration of these \rev{GenAI models} requires robust, scalable computing infrastructures and sophisticated data storage systems, designed to support the relentless and intensive data streams inherent to smart urban environments \citep{wang2023overview}. 
\item[Evaluation of AI-Generated Data:] The quality of AIGC is influenced by several complex factors, including the absence of objective evaluation criteria, challenges in detecting mode collapse, the lack of benchmark data, and issues arising from imbalanced data \citep{saxena2021generative}. Furthermore, there is a notable lack of consensus regarding assessment indicators for various types of \rev{AIGC}, such as time series and spatial data \rev{\citep{lin2023generative, debnath2021exploring}}. This ambiguity in evaluation standards can be attributed partly to the limited number of published studies in this area, which restricts the development of universally accepted benchmarks and metrics \citep{bandi2023power}.
\end{description}

\rev{Based on the Moore's Law and Beyond, continued improvements in semiconductor technology and the advent of novel computing architectures (quantum computing, neuromorphic computing) are expected to significantly enhance computational power \citep{hiremane2005moore}. The recent development of AI-specific hardware, such as GPUs, TPUs, and custom AI accelerators, optimizes the performance of AI models and simulations, making the large-scale integration of GenAI and digital twins more feasible and affordable \citep{mody2016long}}.

\subsection{\rev{Ethical and Societal Challenges} }
\rev{ Building Urban digital twins and smart cities often involve a range of sociotechnical aspects that encompass both the social and technical dimensions of these technologies. From the social perspective, these dimensions include governance and policy, ethical considerations, social equity, and cultural and behavioral changes. Within these applications, GenAI models pose significant ethical and societal challenges, especially in terms of bias and fairness, as highlighted by \cite{ferrara2024genai}. Additionally, these models raise concerns about transparency and accountability.\citep{kenthapadi2023generative}. 
} 

\subsubsection{\rev{Bias and Fairness} } 
\rev{GenAI models, such as GAN and GPT, are trained on large datasets that may reflect the biases present in the real world \citep{grossman2023gptjudge}. These biases can manifest in various ways, leading to unfair or discriminatory outcomes. Addressing bias in AI is essential for maintaining public trust and acceptance \citep{schwartz2022towards}. Bias in AI models primarily originates from two sources: training data and algorithmic design. Training data can carry societal biases related to race, gender, age, and other factors, which the AI model can then replicate and even amplify \citep{mehrabi2021survey}. In the context of urban digital twins, societal biases. }
 
\rev{Even with unbiased data, algorithms themselves can introduce bias through their design and implementation. The impacts of these biases are profound, leading to discriminatory outcomes where certain groups are favored over others, resulting in unfair treatment \citep{ferrara2023fairness}. This not only affects individuals adversely but also erodes public trust in AI technologies, potentially hindering their broader adoption. To mitigate these biases, several strategies are essential. Using diverse datasets that accurately represent various demographics and target user groups can help reduce inherent biases \citep{nazer2023bias}. }
 
 \rev{Employing bias detection tools during the development and deployment phases allows for the identification and mitigation of biases early on. Additionally, conducting regular audits of AI systems ensures ongoing fairness and the ability to address emerging biases effectively. These steps are critical for creating AI models that are equitable, trustworthy, and widely accepted.}

\subsubsection{\rev{Transparency and Accountability in AI Systems} } 
\rev{The complexity and opacity of AI models, often referred to as the "black-box" nature, pose significant challenges for understanding and explaining their decisions \citep{von2021transparency}. Ensuring transparency and accountability is crucial for building trust and meeting regulatory requirements. Black-box models suffer from a lack of interpretability, making it difficult for humans to understand and explain their decision-making processes \citep{hassija2024interpreting}. This lack of transparency can lead to issues with regulatory compliance, as many industries require clear and explainable AI systems. Transparency is essential for building stakeholder trust, as users and stakeholders are more likely to trust AI systems that are clear and understandable \citep{balasubramaniam2022transparency}. Additionally, transparency is crucial for the ethical use of AI, ensuring that these systems are used responsibly and without causing harm.} 

\rev{To ensure accountability, the development of explainable AI (XAI) is vital, as it provides explanations for AI decisions, enhancing transparency \citep{arrieta2020explainable}. Thorough documentation of AI systems, including their development, training data, and decision-making processes, is also necessary. Furthermore, establishing governance frameworks to oversee the ethical use and accountability of AI systems is essential for maintaining transparency and trust in AI technologies \citep{diaz2023connecting}.}

%% file: Content/50-conclusion.tex
\section{Conclusion} \label{sec:Discussion}
This survey investigates and summarizes the recent progress on popular \rev{GenAI} models
and their applications in the smart city domain. To enhance the reliability, performance, and efficiency of urban planning and management, the augmentation and synthesis of urban data based on previously collected and simulated data receive a lot of attention in the transportation, urban mobility, energy systems, infrastructure, and urban \rev{planning} sectors. \rev{GenAI} models support these sectors' urban research and management efforts through their excel capabilities for data augmentation, synthetic data and scenario generation, 3D city modeling, and generative urban design. With the advent of urban digital twin technologies, we envision an integration between the \rev{GenAI} models and cognitive digital twins for a more reliable, autonomous, scalable, cost-effective, and \rev{participatory} management and digital transformation of complex urban systems. Based on a structured review of the previous GenAI application in the smart city sector, this survey outlines \rev{3} promising \rev{directions} for the proposed \rev{GenAI-}enabled urban digital twins and outlines 3 technical strategies to facilitate the \rev{GenAI}-digital twin integration within different software engineering architectures. Finally, the survey closes with a discussion of \rev{potential challenges} that may present in the GenAI-digital twin integration processes \rev{in the context of computing and data science limitations, and ethical and societal concerns}. We believe the integration of GenAI models could significantly revolutionize the existing urban digital twin applications towards more intelligent, reliable, and autonomous systems for supporting smart city development and operations.